\documentclass[table]{article}
\usepackage{iclr2025_conference,times}
\usepackage{xcolor}
\usepackage{amssymb,mathrsfs,amsmath}

\usepackage{graphicx}
\usepackage{subcaption}
\usepackage{wrapfig}
\usepackage{algorithm}
\usepackage{algorithmic}

\usepackage{multirow}
\usepackage{graphicx}
\usepackage[utf8]{inputenc} 
\usepackage[T1]{fontenc}    
\usepackage{hyperref}       
\usepackage{url}            
\usepackage{booktabs}       
\usepackage{amsfonts}       
\usepackage{nicefrac}       
\usepackage{microtype}      
\usepackage{arydshln}

\usepackage{fancyhdr}
\usepackage{lipsum}
\usepackage{geometry}
\usepackage{float}

\usepackage{CJKutf8}

\definecolor{lowyellow}{RGB}{241, 196, 15}

\fancypagestyle{firstpage}{
    \fancyhf{}
    \fancyhead[L]{%
        \hspace{-2mm}
        \includegraphics[height=1cm]{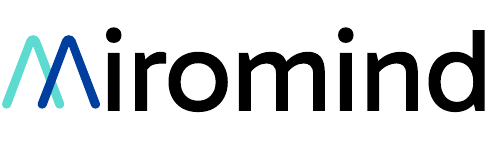} 
        \vspace{1mm}
    }
}

\definecolor{custombluedark}{HTML}{0040A1}
\definecolor{custombluelight}{HTML}{5EDDD2}

\definecolor{mypink3}{cmyk}{0, 0.7808, 0.4429, 0.1412}

\title{MiroMind-M1: An Open-Source Advancement in Mathematical Reasoning via Context-Aware Multi-Stage Policy Optimization}

\author{%
  Xingxuan Li\thanks{\; Equal contribution.}~~,~
  Yao Xiao\footnotemark[1]~~,~
  Dianwen Ng\footnotemark[1]~~,~
  Hai Ye\footnotemark[1]~~,~
  Yue Deng\footnotemark[1]~~,~
  Xiang Lin\footnotemark[1]~~,~
  Bin Wang\footnotemark[1]\\
  \textbf{Zhanfeng Mo},~
  \textbf{Chong Zhang},~
  \textbf{Yueyi Zhang},~
  \textbf{Zonglin Yang},~
  \textbf{Ruilin Li},
  \textbf{Lei Lei}\\
  \textbf{Shihao Xu},~
  \textbf{Han Zhao},~
  \textbf{Weiling Chen},~
  \textbf{Feng Ji},~
  \textbf{Lidong Bing}\thanks{\; Corresponding author: Lidong Bing <lidong.bing@miromind.ai>}\\\\
  MiroMind AI\\\\
  \includegraphics[height=10pt]{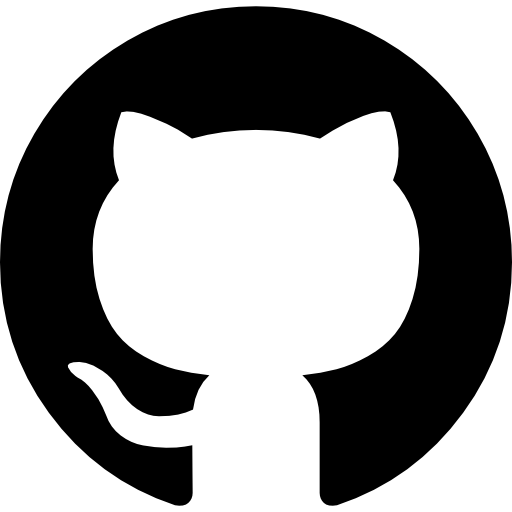} \href{https://github.com/MiroMindAsia/MiroMind-M1}{\texttt{{https://github.com/MiroMindAsia/MiroMind-M1}}}\\
  \includegraphics[height=10pt]{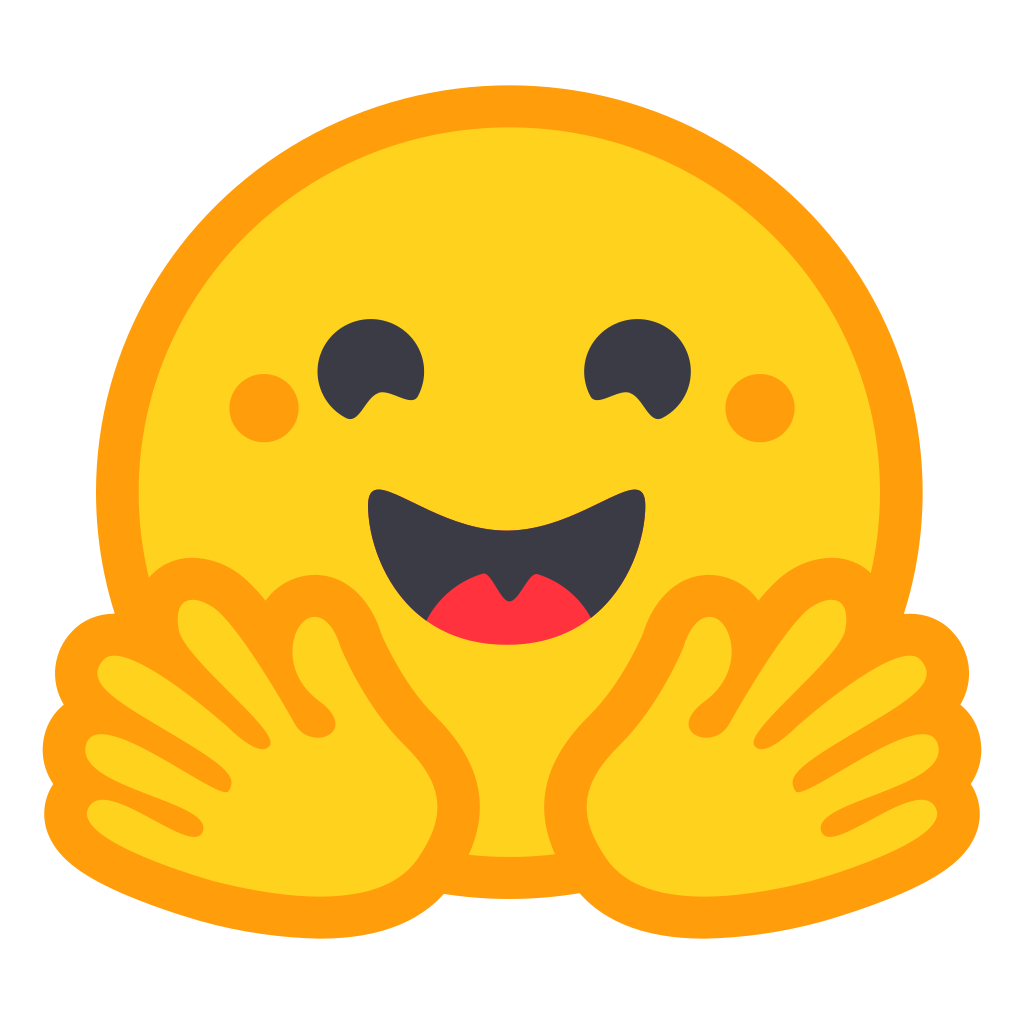} \href{https://huggingface.co/miromind-ai/MiroMind-M1-RL-7B}{\texttt{{https://huggingface.co/miromind-ai/MiroMind-M1-RL-7B}}}\\
  \includegraphics[height=10pt]{figures/hf.png} \href{https://huggingface.co/datasets/miromind-ai/MiroMind-M1-RL-7B}{\texttt{{https://huggingface.co/datasets/miromind-ai/MiroMind-M1-RL-62K}}}\\
  \includegraphics[height=10pt]{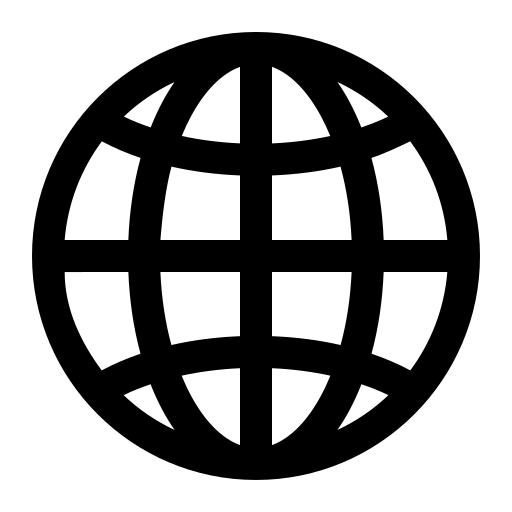} \href{https://miromind.ai/}{\texttt{{https://miromind.ai/}}}
}

\iclrfinalcopy
\begin{document}

\maketitle

\thispagestyle{firstpage}



\begin{abstract}

Large language models have recently evolved from fluent text generation to advanced reasoning across diverse domains, giving rise to reasoning language models (RLMs). Among these domains, mathematical reasoning serves as a representative benchmark as it requires precise multi-step logic and abstract reasoning, which can be generalized to other tasks.
While closed-source RLMs such as GPT-o3 and Claude Sonnet 4 demonstrate impressive reasoning capabilities, their proprietary nature limits transparency and reproducibility. Although many open-source projects aim to close this gap, most of them lack sufficient openness by omitting critical resources such as curated datasets and detailed training configurations, which hinders reproducibility.
To contribute toward greater transparency in RLM development, we introduce the \textbf{\texttt{MiroMind-M1}} series, a set of fully open-source RLMs built on the Qwen-2.5 backbone that match or exceed the performance of existing open-source RLMs.
Specifically, our models are trained in two stages: supervised fine-tuning (SFT) on a carefully curated corpus of 719K math-reasoning problems with verified chain-of-thought (CoT) trajectories, followed by reinforcement learning with verifiable reward (RLVR) on 62K challenging and verifiable problems. To enhance the robustness and efficiency of the RLVR process, we introduce Context-Aware Multi-Stage Policy Optimization (CAMPO), an algorithm that integrates length-progressive training with an adaptive repetition penalty to encourage context-aware RL training.
Our model achieves state-of-the-art or competitive performance and superior token efficiency among Qwen-2.5-based open-source 7B and 32B models on the AIME24, AIME25, and MATH benchmarks.
To facilitate reproducibility, we release the complete stack: models (\texttt{MiroMind-M1-SFT-7B}, \texttt{MiroMind-M1-RL-7B}, \texttt{MiroMind-M1-RL-32B}); datasets (\texttt{MiroMind-M1-SFT-719K}, \texttt{MiroMind-M1-RL-62K}); and all training and evaluation configurations.
We hope these resources will support further research and foster community advancement.

\end{abstract}

\newpage

\tableofcontents
\clearpage

\section{Introduction}





Recent advances in large language models (LLMs), driven by the Transformer architecture \citep{attention_is_all_you_need}, massive pre-training \citep{Scaling_Laws_for_LM,gemini2.5pro,dubey2024llama3herdmodels}, and emergent abilities such as in-context learning \citep{GPT_llm_are_fewshot_learners,dong-etal-2024-in-context-learning-survey}, have led to strong performance in planning, reasoning, and problem solving \citep{valmeekam2023on_planning_abilities_of_llm,feng2025reasoning,xu2025largereasoningmodelssurvey,yao2023react}. Among these capabilities, structured multi-step reasoning has captured significant interest, shifting the focus from fluent generation to cognitively grounded chain-of-though (CoT) reasoning \citep{wei2022finetuned,CoT_JasonWei}. 
Building on this momentum, researchers have developed reasoning language models (RLMs), specialized LLMs explicitly trained to produce multi-step CoT for complex reasoning tasks. 
This approach has led to significant performance improvements across various domains, including mathematics \citep{GRPO,deepscaler2025}, question answering \citep{wei2024measuringshortformfactualitylarge,wei2024longform_factuality_QA}, planning \citep{wei2025plangenllmsmodernsurveyllm}, and code generation \citep{deepcoder2025,code-r1,xu2025kodcodediversechallengingverifiable}. Specifically, many studies \citep{GRPO,deepscaler2025,skywork-or1-2025,OpenReasonerZero2025,lightr1} focus on mathematical reasoning as it demands robust multi-step logical thinking and abstract problem-solving skills, which are fundamental to advanced reasoning in language models.
While RLMs have achieved unprecedented performance, recent progress has been dominated by a handful of proprietary, closed-source models \citep{openai2024gpt4o,claude3,gemini2.5pro}, the training data, methodologies, and evaluation protocols of these commercial systems remain largely unclear, creating significant barriers to transparency, reproducibility, and further scientific innovation. As a result, the research community is urgently seeking open and reproducible pathways for building high-performing RLMs. 
In response, a series of open-source initiatives have emerged, demonstrating that state-of-the-art reasoning capabilities can be achieved through a two-stage training paradigm: supervised fine-tuning (SFT) on curated reasoning traces, followed by reinforcement learning with verifiable rewards (RLVR)~\citep{Deepseek_R1, KIMI_scaling_RL, qwen3, mistralai2025magistral}. In the SFT stage, the pretrained model is fine-tuned on curated datasets of annotated QA pairs and verified CoT traces, enabling it to learn step-by-step reasoning through imitation. In the subsequent RLVR stage, the model is further optimized on reasoning tasks using reinforcement learning to maximize a task-specific reward signal. This process encourages the exploration of more effective and robust CoT patterns. Rewards are typically predefined and computed by an external verifier, with higher scores assigned to correct answers and lower scores to incorrect ones. 

However, despite the overall training paradigm being disclosed by these open-source projects, reproducing high-performing reasoning models in practice remains challenging due to missing or insufficiently documented implementation details. In the SFT stage, key components such as data collection and curation strategies, training data composition, curriculum design, and configuration details are often under-specified or ambiguously described. Similarly, in the RLVR stage, key aspects such as sampling strategies, data composition, and reward function implementations are often unclear or not publicly disclosed. The lack of transparency severely limits reproducibility and hinders further progress on improving reasoning capabilities. 



To promote greater transparency in RLM development and advance cutting-edge research, we present a comprehensive and well-documented reproduction of reinforcement-incentivized RLMs. We specifically focus on training RLMs for mathematical reasoning, a representative and practical testbed to evaluate reasoning capabilities, where strong performance reliably reflects the ability of an LLM to generate logically consistent multi-step CoTs. As each LLM-generated answer can be rigorously verified to provide a clear reward signal, this domain presents an ideal setting for RLVR. Moreover, the abundance of open-source, well-annotated mathematical reasoning datasets greatly facilitates reproducibility.


This report details our reproduction of RLMs, covering key aspects such as training dataset curation, open-source codebases, and the implementation and evaluation of both SFT and RLVR. We begin by carefully curating a high-quality SFT dataset, drawing from publicly available sources such as OpenR1~\citep{openr1}, OpenThoughts~\citep{openthoughts}, Light-R1~\citep{lightr1}, and Synthetic-1~\citep{2025synthetic1}.
To ensure data quality and enable fair comparisons, we perform rigorous deduplication and decontamination, removing duplicate samples and filtering out any overlap with evaluation benchmarks such as AIME24, AIME25, and MATH500~\citep{AoPSAIME,lightman2023lets}.
Through systematic ablation studies on the impact of dataset properties on SFT performance, we find that incorporating longer and more complex reasoning trajectories consistently leads to substantial improvements. 
This highlights the importance of trajectory depth and semantic richness in CoT reasoning traces for enhancing mathematical reasoning through SFT.
Motivated by this, we propose CAMPO, a context-aware RLVR algorithm that promotes progressively longer CoT reasoning traces via a length-based curriculum. 
Specifically, CAMPO employs a multi-stage training strategy with progressively increasing context lengths and a dynamic repetition penalty to reduce redundancy and improve training stability.
Furthermore, we open-source our rigorous data curation pipeline for RLVR and release an improved, more robust math verifier to support a more stable and effective RLVR training process. Leveraging the Qwen2.5 base models~\citep{qwen2.5}, we develop the \texttt{MiroMind-M1} series of mathematical RLMs trained on carefully curated datasets through a two-stage process of SFT and RLVR, employing CAMPO during the RLVR stage.
Comprehensive evaluations show that our \texttt{MiroMind-M1-RL} achieves state-of-the-art or competitive performance and token efficiency among Qwen2.5-based open-source models on mathematical benchmarks such as AIME24, AIME25, and MATH500. 
In summary, our main contributions are listed as follows:


\begin{itemize}
    \item We release a high-quality dataset combining complicated mathematical reasoning problems (\texttt{MiroMind-M1-RL-62K}) and long-form reasoning trajectories (\texttt{MiroMind-M1-SFT-719K}) constructed based on openly available data sources.
    \item We introduce CAMPO, a context-aware reinforcement learning framework with staged context expansion and repetition penalties for training high-performing RLMs.
    \item We present the MiroMind series, including \texttt{MiroMind-SFT-7B}, a strong supervised baseline, and \texttt{MiroMind-M1-RL}, which achieves state-of-the-art or competitive performance among fully open-sourced 7B or 32B math reasoning models. 
    \item We open-source the full stack, datasets, models, code, and an improved robust verifier, to enable reproducible math reasoning research on mathematical RLMs. We share key insights and lessons learned from data curation and the SFT-RLVR process to advance future research in RLMs.
\end{itemize}





\section{Related Work}

\subsection{Reasoning Language Models}
RLMs have become a promising direction for advancing large language models. Several leading proprietary models,
such as OpenAI’s o-series~\citep{openai-o3}, Anthropic’s Claude~\citep{claude3}, Gemini 2.5 Pro~\citep{gemini2.5pro}, Seed \citep{seed-thinking-1.5}, and Magistral \citep{mistralai2025magistral}, place a strong emphasis on structured multi step reasoning.
Recent models including DeepSeek-R1~(e.g., R1, R1-0528)~\citep{Deepseek_R1}, Qwen~(QwQ, Qwen3)~\citep{qwq-32b,qwen3}, and Gemma~\citep{gemma} have partially disclosed their training pipelines, providing insight into the emerging paradigm for reasoning enhancement.

Mathematics has become the primary benchmark for evaluating RLMs, due to its objective verifiability and controllable difficulty. High-level benchmarks such as AIME24, AIME25, and MATH500 remain unsolved and have become central to recent progress. To tackle these challenges, most RLMs follow a two-stage training pipeline: SFT on curated datasets with verified CoT traces, followed by reinforcement learning RLVR.

Although methods such as GRPO~\citep{GRPO} and reward-based optimization have been adopted in models like DeepSeek-R1 and Qwen, many training details (e.g., data composition, reward design, and sampling strategies) remain undisclosed. This has prompted numerous open-source replication efforts (e.g., AM~\citep{AM}, Light-R1~\citep{lightr1}, Open-R1~\citep{openr1}) to explore and refine best practices. 
Notably, general models trained with SFT and RLVR often outperform those trained primarily on math (e.g., Qwen2.5-Math), underscoring the value of verifiable feedback and broad reasoning supervision.


\subsection{Supervised Fine-Tuning}

The success of DeepSeek distilled models \citep{Deepseek_R1} demonstrates that SFT on massive data can equip LLMs with advanced reasoning capabilities. 
On the one hand, many replication works make efforts to collect large-scale supervised datasets from publicly available resources of various domains for fine-tuning reasoning language models.
The OpenThoughts datasets \citep{guha2025openthoughtsdatarecipesreasoning} curates a synthetic reasoning dataset with 114k examples, covering multiple domains including math, science, coding, and puzzles. The CoTs are generated by DeepSeek-R1 and verified.
They further conduct comprehensive ablations to find the best question resources and scale up their dataset to 1.2M with verified CoTs from QwQ-32B. 
AM-DeepSeek-R1-Distilled-1.4M \citep{zhao202514millionopensourcedistilled} is an another compelling dataset with 1.4M samples covering math, code, science, general chat, instruction following and dialogue. Among these, 500k samples are cleaned from other open-source datasets, and the other 900k samples are distilled from the DeepSeek-R1-671B, filtered based on quality and assigned with difficulty levels. 
Mixture-of-Thoughts \citep{openr1} collects 350k verified reasoning traces distilled from DeepSeek-R1, spanning tasks in math, code, and science. This dataset is known as designed to teach language models to reason step-by-step.
Synthetic-1 \citep{2025synthetic1} constructs 894k samples distilled from DeepSeek-R1, covering math, code and STEM domains.
All these works emphasized the importance of data cleaning, deduplication and verification. 
For verification, besides from rule-based verifiers for math problems and unit-test-based validation for code problems, these works also use LLM-as-judge \citep{guha2025openthoughtsdatarecipesreasoning,2025synthetic1} and reward model evaluation \citep{zhao202514millionopensourcedistilled} to verify the samples from general domains. 

On the other hand, several open-sourced models have already achieved strong performance on verifiable reasoning tasks, especially math problem solving. 
OpenR1-Qwen-7B \citep{openr1} leverages 220k math problems with high quality long CoTs to achieve comparable performance towards the corresponding DeepSeek distilled model.   
Light-R1 \citep{wen2025light} constructs a two-stage SFT training pipeline to boost its performance. 
In the first stage, the model leverages around 76k difficult and diverse samples with verified long-form CoT responses to reproduce a similar training procedure to DeepSeek distilled models. 
In the second stage, the model leverages a filtered subset of 3k samples which are not completely tackled by DeepSeek-R1-Distill-Qwen-32B and DeepSeek-R1 to further improve model performance. 
AM-Thinking-v1 \citep{ji2025amthinkingv1advancingfrontierreasoning} takes advantage of larger learning rate and batch size during SFT to get rid of SFT pattern shifts \citep{tian2025deepdistillenhancingllmreasoning}. 
These advancements highlight a promising trend that open-source models can excel in complex reasoning tasks by carefully configuring on data curation and training procedure.

\subsection{Reinforcement Learning with Verfiable Rewards}

RLVR has emerged as the most promising paradigm for advancing the reasoning abilities of language models, distinguished by its comprehensive approach to data quality, algorithmic innovation, reward design, and training methodology~\citep{zhang2025100daysdeepseekr1survey}. 
At the core of RLVR are high-quality and verifiable datasets,
including DeepScaleR~\citep{deepscaler2025}, Skywork OR1~\citep{skywork-or1-2025}, Open Reasoner Zero~\citep{OpenReasonerZero2025}, DeepMath~\citep{he2025deepmath103klargescalechallengingdecontaminated}, among others. These datasets are meticulously curated to ensure a suitable balance of medium difficulty, thorough removal of benchmark leakage, comprehensive cross-referencing of answers, and exclusion of uncertain or ambiguous cases. Such rigorous data preparation guarantees both verifiability and maximal learning effectiveness.

On the algorithmic side, RLVR has progressed beyond foundational methods such as PPO~\citep{PPO} and GRPO~\citep{GRPO}, incorporating specialized variants including DAPO~\citep{DAPO}, Dr. GRPO~\citep{DrGRPO}, VC-PPO~\citep{VC-PPO}, VAPO~\citep{yuyue2025vapoefficientreliablereinforcement-vapo}, and others. These algorithms are crafted to address specific challenges like length bias, difficulty imbalance, and value function management for long reasoning chains, while also promoting training stability. In addition, reward engineering plays a critical role by providing clear and verifiable signals, such as correctness, adherence to answer formats, and penalties for inefficient or excessively long responses~\citep{zhang2025100daysdeepseekr1survey}. These mechanisms help to minimize reward hacking and foster robust, stepwise reasoning.

With respect to training strategies, RLVR adopts curriculum learning by gradually increasing the complexity or length of tasks, which supports the systematic acquisition of reasoning skills~\citep{zhang2025100daysdeepseekr1survey}. Techniques such as KL loss regularization—sometimes selectively omitted to encourage exploration—and rollout pruning are also employed to discard low-value trajectories~\citep{xu2025rolloutsusefuldownsamplingrollouts}, thereby improving both training efficiency and stability.

By systematically integrating advances in data, algorithms, rewards, and training practices, RLVR provides a clear and scalable path toward reliable and high-performing chain of thought generation in large language models.

\section{Enhancing Math Reasoning Capabilities of Language Models via Supervised Fine-Tuning}
\label{sec:SFT}
In this section, we explore how to construct high-quality SFT data capable of replicating or even surpassing the performance of the DeepSeek-R1-distilled models \citep{Deepseek_R1} for math reasoning. 
We detail methods for sourcing and processing data.
In addition, we share insights and lessons learned throughout the data construction and SFT training process. 

\subsection{Data Curation}
\subsubsection{Data Collection}

We focus mainly on enhancing the math reasoning capabilities of LLMs in this work, so we only collect the R1 distillation data from the math domain.
We collect data from four public sources which are OpenR1~\citep{openr1}, Open-thoughts~\citep{guha2025openthoughtsdatarecipesreasoning}, Light-R1~\citep{wen2025lightr1curriculumsftdpo}, and Synthetic-1~\citep{2025synthetic1}.
We display the detailed data statistics of these datasets in Table~\ref{tab:data-collection}.

\noindent{\textbf{OpenR1.}}
OpenR1~\citep{openr1} sources approximately 400K math problems from AI-MO/NuminaMath-1.5. 
Each question includes between 1 and 8 reasoning traces from DeepSeek-R1. 
We retain all traces with correct final answers, verified by MathVerify~\citep{math_verify} and an LLM-based judge.

\noindent{\textbf{OpenThoughts.}}
Unlike OpenR1, OpenThoughts~\citep{guha2025openthoughtsdatarecipesreasoning} includes questions from diverse domains such as math, science, coding and puzzles.
It contains 114K questions with reasoning traces generated by DeepSeek-R1. 
We retained only math-related questions, resulting in approximately 56K question-response pairs. 

\noindent{\textbf{Light-R1.}}
Light-R1~\citep{lightr1} sources math questions from OpenR1-Math-220K, OpenThoughts-114K, LIMO, OpenMathInstruct-2, s1K, Omni-MATH, Hendrycks\_Math, and AIME (up to 2023).
The dataset is decontaminated against common reasoning benchmarks such as AIME2024, AIME2025, MATH500, and GPQA Diamond.
Reasoning traces are generated using DeepSeek-R1.
The final datasets include 76k and 3k challenging problems for stage 1 and stage 2 training, respectively.


\noindent{\textbf{Synthetic-1.}}
Synthetic-1~\citep{2025synthetic1} is a large-scale dataset for supervised fine-tuning, featuring reasoning traces produced by DeepSeek-R1. 
It covers a range of domains including Math, Algorithmic Coding, Real-World Software Engineering, Open-Ended STEM Q\&A, and Synthetic Code Understanding. 
The math questions are derived from NuminaMath and undergo LLM-driven postprocessing to transform multiple-choice items into free-form questions and remove those without automatically verifiable answers (e.g., proof-based queries). 
For our work, we focus only on math-related problems, which results in 247k problems.


\begin{table}[t]
\centering
\resizebox{0.7\textwidth}{!}{
\begin{tabular}{lcccc}
\toprule
\textbf{Dataset} & \textbf{\# Questions} & \textbf{\# Traces} & \textbf{\# Traces used in SFT} & \textbf{Category}  
\\ \hline
OpenR1           & 191k             & 418k            & 418k                         & math                            \\
Open-thoughts    & \phantom{0}56k               & \phantom{0}56k              & \phantom{0}56k                          & math     \\
Light-R1         & \phantom{0}75k               & \phantom{0}76k              & \phantom{0}76k                          & math      \\
Synthetic-1      & 362k              & 638k             & 247k                     & math                \\
Ours      & 412k              & 719k             & 719k                     & math                \\
\bottomrule                                
\end{tabular}}
\caption{Statistics of math-focused datasets used for SFT, including the number of prompts, total reasoning traces, and traces selected for training.}
\label{tab:data-collection}
\end{table}

\subsubsection{Data Processing}

\paragraph{Data Deduplication}

To ensure data quality, we first remove duplicate question-response pairs by computing $N$-gram overlaps on the concatenated query and response text. 
Samples with significant overlap are excluded to reduce redundancy in the training data. Note that we allow one question to have multiple correct responses. 

\paragraph{Data Decontamination}
\begin{wrapfigure}{r}{0.5\textwidth}
    \centering
    \includegraphics[width=0.48\textwidth, trim=8 10 6 6, clip]{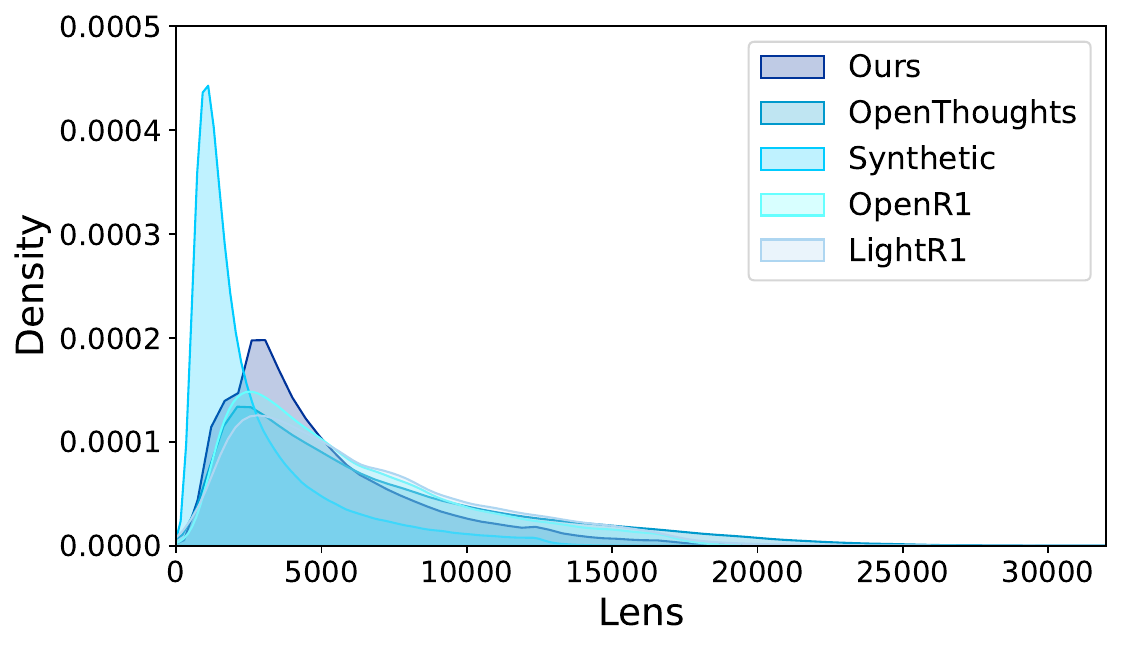}
    \caption{Length distribution for of datasets.}
    \label{fig:sft_len_dist}
    \vspace{-1em}
\end{wrapfigure}
To avoid data contamination and ensure fair evaluation, we perform decontamination for training data with respect to our target evaluation sets. 
Specifically, we apply an $N$-gram overlap filter to remove any training samples whose questions match those in Math500, AIME24, or AIME25. 
This filtering effectively prevents data leakage.

\paragraph{Data Statistics of the Final Training Set} 
We end up with 719K SFT training samples shown in Table~\ref{tab:data-collection}. 
We demonstrate the length distribution of thinking tokens in Figure~\ref{fig:sft_len_dist} , together with Open-R1, Synthetic-1, Open-Thoughts and Light-R1.

\subsection{Supervised Fine-tuning}

\noindent{\textbf{Experimental Setup.}}
\label{sft_setup}
We train our models for 3 epochs using a peak learning rate of $5.0 \times 10^{-5}$ with a cosine learning rate scheduler. 
We set the warmup step ratio to 10\% and use a batch size of 128.
To support long generations of complex reasoning, we increase model's \emph{max\_position\_embeddings} to 32,768 {using Linear RoPE scaling}.
We adopt a no-packing strategy for training because of its empirical advantages.
More details on why we employ no-packing can be found in Section~\ref{why_nopacking}.
Our implementation is based on LlamaFactory~\citep{zheng2024llamafactory}, with a cutoff length of 26,000.
We follow Deepseek-R1~\citep{Deepseek_R1} to employ Qwen-2.5-Math-7B as the initial checkpoint.

\subsubsection{Main Experiment Results}

For evaluation, we report avg@k to ensure stable and reliable results, setting $k=64$ for AIME24 and AIME25, and $k=5$ for MATH-500. 
For evaluation, we use a maximum generation length of 32,768 tokens, a sampling temperature of 0.6, and a top-$p$ value of 0.95.

We report our results in Table~\ref{tab1:sft_performance}.
Our model achieves 60.5 on AIME24, 45 on AIME25, and 94.6 on MATH-500.
Across all three tasks, our model consistently surpasses other SFT models of the same size, demonstrating the quality of our constructed dataset.
In particular, our model also outperforms a recent SFT model, \texttt{MiMo-7B-SFT}~\citep{xiaomi2025mimo}.


\begin{table}[h]
\centering
\resizebox{0.9\textwidth}{!}{%
\begin{tabular}{llccc}
\toprule
\textbf{Project} & \textbf{Initial Checkpoint} & \textbf{AIME24} & \textbf{AIME25} & \textbf{MATH500} \\
\midrule
DeepSeek-R1~\citep{Deepseek_R1} 
& Qwen2.5-Math-7B 
& 55.5  & 40.4\textsuperscript{†} & 92.8  \\

OpenThoughts~\citep{openthoughts} 
& Qwen2.5-7-Instruct
& 31.3
& 23.3\phantom{0}
& 83.2 \\

Open-R1~\citep{openr1} 
& Qwen2.5-Math-7B-Instruct 
& 36.7 & 40.0\phantom{0} & 90.6 \\

Synthetic-1~\citep{2025synthetic1} 
& Qwen-2.5-7B-Instruct
& 30.0 & 26.6\phantom{0} & 85.6 \\

MiMo-7B-SFT~\citep{xiaomi2025mimo}
& MiMo-7B-Base 
& 58.7  & 44.3\phantom{0} & 93.0  \\

\midrule

MiroMind-SFT-7B & Qwen-2.5-Math-7B & 60.4 & 45.0\phantom{0} & 94.6\\

\bottomrule
\end{tabular}
}
\caption{We report results of our model on math reasoning benchmarks. It outperforms the distilled models trained from the same initial checkpoint by Deepseek. \textsuperscript{†} means that the score of DeepSeek-R1 on AIME25 is from our evaluation.}
\label{tab1:sft_performance}
\end{table}

\subsubsection{Insights}
In this section, we show some experiences obtained in the data curation and training process.
We first elaborate on why we use no-packing for training.
We also show that trajectory length is a simple but effective metric for sample selection. 
We finally present a strategy to mix packing and no-packing to improve training efficiency while preserving performance. 
The training of these ablations follows Section~\ref{sft_setup}, unless otherwise specified.
And experiments are based on Qwen2.5-Math-7B-Instruct and default split of OpenR1-Math-220k \footnote{We select Qwen2.5-Math-7B-Instruct because it can achieve better performance than other models(e.g. Qwen-2.5-Math-7B) on Open-R1 dataset.}.

\paragraph{No-packing outperforms packing.}
\label{why_nopacking}

In this part, we first clarify the difference between multiple packing strategies.
Instead of training each sequence individually as no-packing, packing concatenates multiple sequences into a single input to maximize the usage of GPU memories, which significantly improves the training efficiency.
However, this packing strategy might introduce cross-sample attention pollution, where tokens from different sequences within the same packed input may inadvertently attend to each other.
To address this, neat-packing are usually employed during training to ensure that each token can only attend to tokens within its own original sequence.
We subsequently compared the performance of these strategies.
We report results in Table~\ref{tab:packing}.
We observe that no-packing outperforms packing and neat-packing \footnote{Note that these results are based on LlamaFactory's implementation, which uses a knapsack algorithm to minimize padding by packing as many sequences as possible. This approach may violate the i.i.d. assumption in training, potentially contributing to the observed inferior performance of the packing and neat-packing strategies.}. 

\begin{table}[h]
\centering
\begin{minipage}{0.45\textwidth}
\raggedright
\resizebox{\textwidth}{!}{%
\begin{tabular}{lccc}
\toprule
\textbf{Strategy} & \textbf{AIME24} & \textbf{AIME25} & \textbf{MATH500} \\
\midrule
Packing      & 35.41 & 26.66 & 89.06 \\
Neat-packing & 32.50    & 26.25    & 88.80    \\
No-packing   & \textbf{38.12}    & \textbf{29.37}    & \textbf{90.40}    \\
\bottomrule
\end{tabular}
}
\caption{No-Packing usually leads to better performance.}
\label{tab:packing}
\end{minipage}%
\hfill
\begin{minipage}{0.48\textwidth}
\raggedright
\resizebox{\textwidth}{!}{%
\begin{tabular}{clccc}
\toprule
\textbf{\# Data} & \textbf{Method} & \textbf{AIME24} & \textbf{AIME25} & \textbf{MATH500} \\
\midrule
\multirow{2}{*}{30k} & Random & 29.58 & 23.12 & 86.93 \\
                    & Long   & \textbf{35.21} & \textbf{24.79} & \textbf{88.66} \\
\midrule
\multirow{2}{*}{50k} & Random & 31.66 & 24.58 & \textbf{89.93} \\
                    & Long   & \textbf{35.41} & \textbf{30.21} & 89.80  \\
\bottomrule
\end{tabular}
}
\caption{Training on the data points that have long trajectories obtains better performance.}
\label{tab:dataselection}
\end{minipage}
\end{table}

\paragraph{Response length is a strong metric for sample selection.}
In the data preparation process, we find that training on longer trajectories usually leads to better performance.
To support the finding, we select a subset of data samples from the full dataset for training here. 
We primarily consider two selection strategies: \emph{random} and \emph{long}. 
For the \emph{random} approach, we randomly sample data points from the dataset. 
For the \emph{long} approach, we select the longest samples. 
In both cases, we ensure that the number of selected samples remains identical.
We find that \emph{long} consistently outperforms \emph{random} with different sample scales. 
We report results in Table~\ref{tab:dataselection}.
In almost all cases, models training on samples selected by \emph{long} outperform those of \emph{random}.
The reason may be that longer trajectories are more likely to arise from complex questions that can better satisfy challenging math tasks such as AIME.

\paragraph{Packing followed by no-packing can save time without significant performance drop.}
Although the no-packing strategy can produce better performance, it is less efficient compared to the other two methods. 
To balance performance and training efficiency, we attempt to first train our models with packing for the first two epochs, and then switch to no-packing for the final epoch.
The performance on math 500 is 90.4 and 89.6, respectively.
This approach allows us to leverage the benefits of no-packing while reducing the overall training time.

\section{Boosting Reasoning Performance and Efficiency with Reinforcement Learning}

\begin{figure}[htb]
    \begin{minipage}[t]{.4\textwidth}
        \centering
        \includegraphics[width=\textwidth]{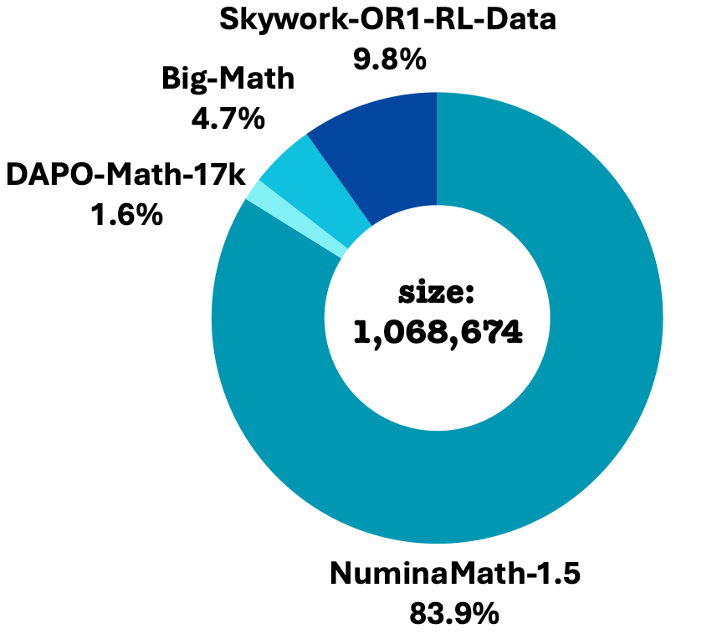}
        \caption{Initial composition distribution. Big-Math comprises HARP and reformulated machine outputs \citep{albalak2025bigmathlargescalehighqualitymath}; Skywork-OR1-RL-Data \citep{skywork-or1-2025} contains only maths.}\label{fig:1}
    \end{minipage}
    \hfill
    \begin{minipage}[t]{.51\textwidth}
        \centering
        \includegraphics[width=\textwidth]{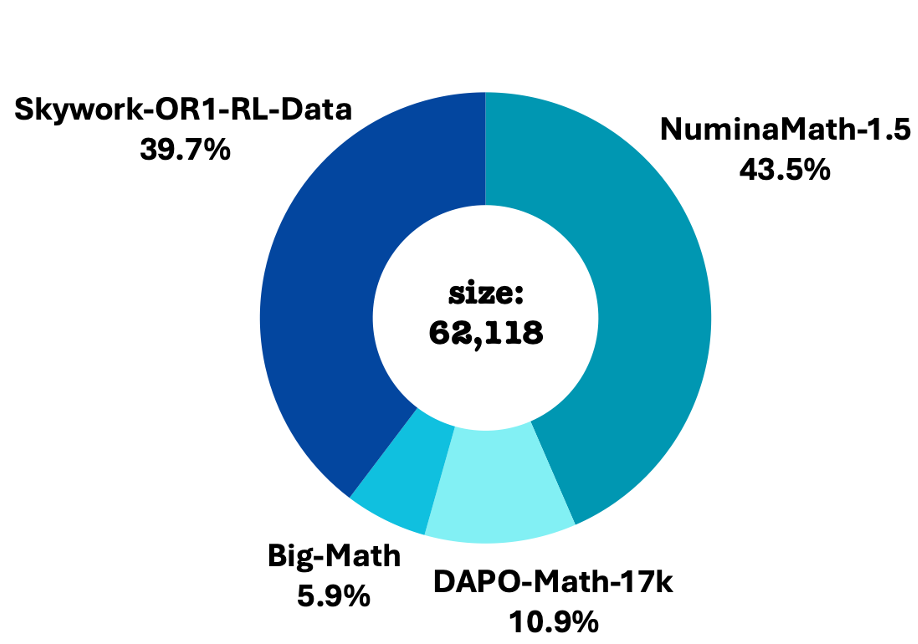}
        \caption{Final composition distribution. Data are filtered using the inclusion-exclusion criteria described in Section \ref{sec:rl_data} and the illustration shown in Figure \ref{fig:RL-datasource-selection}. }\label{fig:2}
    \end{minipage}  
    \label{fig:1-2}
\end{figure}

Recent studies have demonstrated that RL can enhance reasoning performance in both base \citep{DAPO, wen2025light} and R1-distilled LLMs \citep{deepscaler2025, skywork-or1-2025}.
A key observation is that performance gains are often accompanied by increased response length.
However, longer output may introduce redundancy and unnecessary repetition \citep{xie2025logicrlunleashingllmreasoning}, suggesting that efficiency is an equally important yet under-explored aspect.
In this section, we introduce the \texttt{MiroMind-M1-RL} model series, including \texttt{MiroMind-M1-RL-32B} and \texttt{MiroMind-M1-RL-7B}.
\texttt{MiroMind-M1-RL} not only improves performance but also enhances efficiency in mathematical reasoning.
We begin by describing the data preparation process in Section \ref{sec:rl_data},  including the selection criteria. 
Section \ref{sec:rl_pre} provides preliminaries on RL-based training.
Section \ref{sec:rl_campo} introduces the context-aware multi-stage policy optimization (CAMPO) algorithm used to train the \texttt{MiroMind-M1-RL} model series, along with an analysis of each of its components.
Section \ref{sec:rl_32b} and \ref{sec:rl_7b} present the \texttt{MiroMind-M1-RL-32B} and \texttt{MiroMind-M1-RL-7B} models, respectively.

\subsection{Data Collection}
\label{sec:datapre}
To enhance mathematical reasoning capabilities, we incorporate data from diverse sources to ensure broader coverage and increased complexity of mathematical topics. Our dataset includes NuminaMath-1.5 \citep{numina_math_datasets} with 896K problems; Skywork-OR1-RL-Data \citep{skywork-or1-2025}, containing a 105K-size math subset; Big-Math \citep{albalak2025bigmathlargescalehighqualitymath}, comprising 50K problems from both reformulated content and HARP \citep{yue2024harp}; and DAPO-Math-17K \citep{DAPO}, a carefully curated collection of 17K high-quality math problems. These datasets constitute the initial selection of our training candidates, forming a total pool of approximately 1M examples, as illustrated in Figure \ref{fig:1}. 

\label{sec:rl_data}

\begin{figure}
    \centering
    \includegraphics[width=0.78\linewidth]{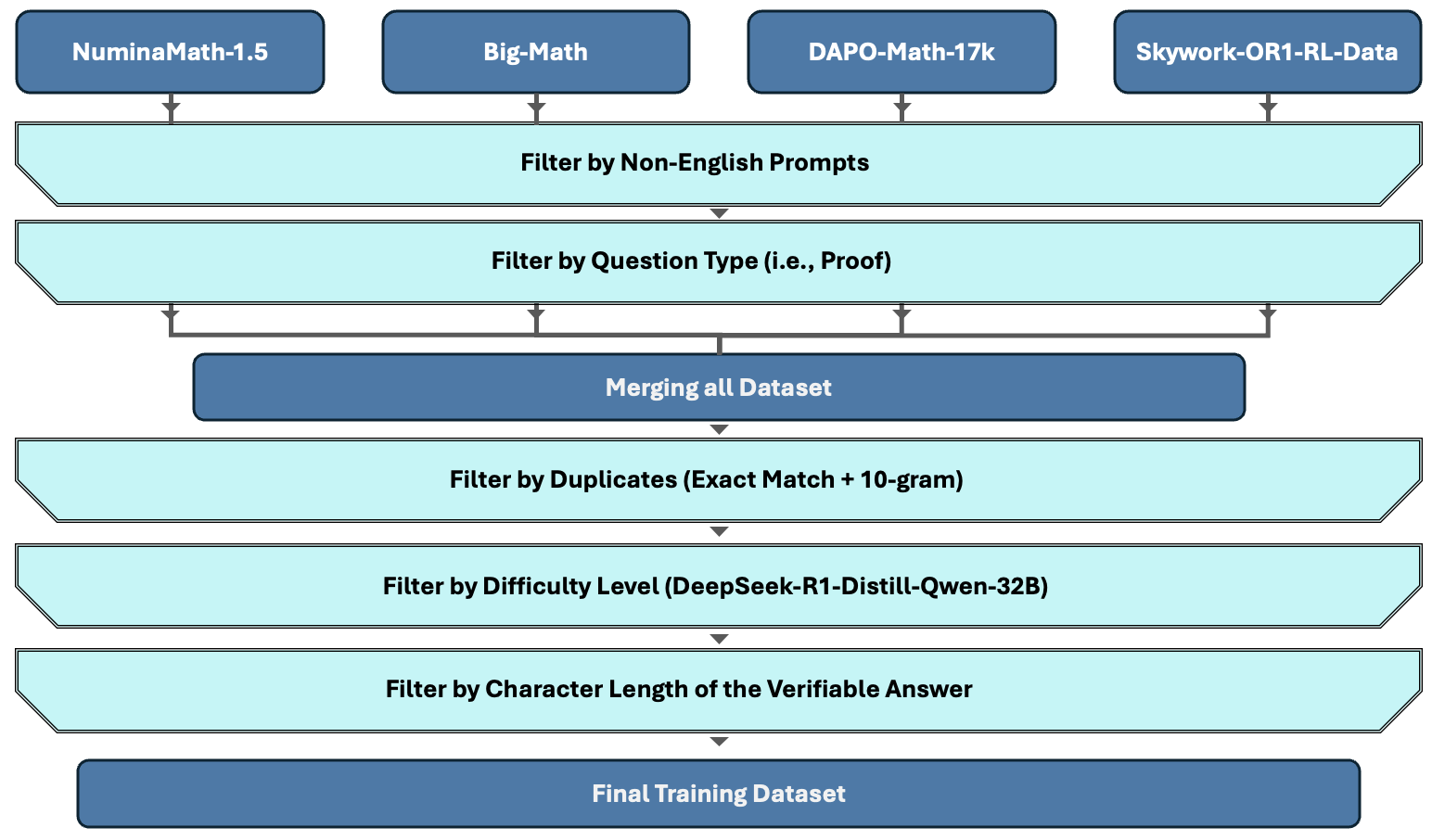}
    \caption{Inclusion-Exclusion Criteria. Overview of the filtering strategy used to construct the final training dataset, consisting of 62,118 problems selected from an initial pool of 1M candidates drawn from four different data sources. The filtering process, which applies criteria such as de-duplication, difficulty-based pruning, and verifiability constraints, results in the exclusion of approximately 94\% of the original data. 
    }
    \label{fig:RL-datasource-selection}
\end{figure}

Despite the dataset’s broad coverage of mathematical topics, rigorous curation is necessary to maintain the quality standards needed for successful reinforcement learning. Some problems, such as those requiring long, free-form answers or detailed mathematical proofs, are difficult for rule-based verifiers to evaluate accurately. In these cases, even correct responses may be mistakenly marked as incorrect due to limitations in the verification logic. Other problems involve ambiguous or incomplete answer formats, especially when the final answers are lengthy or contain multiple valid interpretations. These cases may result in inconsistent labeling, where partially correct answers are penalized despite capturing the key idea. Such misjudgments introduce conflicting signals during training. That is, the model may receive negative feedback for answers that are actually valid, or positive feedback for flawed ones. 
These misleading updates not only hinder the model's ability to generalize but also make learning unstable and, in some cases, cause the model to collapse.
To mitigate these issues, we perform a series of filtering steps to refine the candidate set, as part of the overall data processing pipeline outlined in Figure~\ref{fig:RL-datasource-selection} and explaining the rationale below, which depicts our selection criteria.
\begin{enumerate}
    \item \textbf{Filter by style/format:} We begin by filtering the collected math problems based on style and format, removing proof-based questions that are non-verifiable. These tend to be open-ended and too complex for consistent evaluation using rule-based methods. To ensure linguistic consistency, we also restrict the dataset to English-language problems, as our current study does not explore the effects of multilingual learning. 
    
    \item \textbf{Filter by duplicates:} To reduce redundancy and improve the diversity of training signals, we perform deduplication in two stages. First, we remove math problems that are exact duplicates. Second, we eliminate near-duplicates by applying a 10-gram similarity threshold, which helps filter out semantically similar problems that may differ only in surface form. This ensures a cleaner dataset that better supports generalization during training.

    \item \textbf{Filter by difficulty:} Prior studies \citep{DAPO, skywork-or1-2025, wen2025light} have shown that not all training examples are equally beneficial for learning. This challenge becomes more prominent in the GRPO framework, where batches consisting entirely of correct or incorrect responses result in zero advantage, leading to uninformative gradient updates.
    To mitigate this, we conduct an offline difficulty-aware filtering strategy that adapts to the model’s current performance. For each math problem, we generate five rollouts using \texttt{Deepseek-R1-Distill-Qwen-32B} (with temperature set to 1.0 and a maximum output length of 32K tokens). We exclude extreme cases, such as those with fully correct or fully incorrect responses, and instead focus on problems that are neither trivial nor unsolvable. This targeted selection increases the likelihood of generating informative learning signals during RL training. Additionally, we find that including relatively simple problems (e.g., those with a pass rate of 0.8) is beneficial. These examples reinforce the model’s existing knowledge and help prevent divergence during training. This stabilizing effect is particularly valuable when KL regularization is not applied, as such problems act as anchors that support steady learning. In contrast, always presenting harder problems with lower success rates may overly emphasize exploration, potentially leading the model away from already-correct reasoning patterns. For datasets such as Skywork-OR1-RL,
    which include offline difficulty estimates for each problem relative to the performance of \texttt{Deepseek-R1-Distill-Qwen-32B} (measured over 16 independent rollouts), we also exclude both ends of the spectrum, removing trivial (fully correct) and unsolvable (fully incorrect) cases from the training examples.
    
    \item \textbf{Filter by character length of verifiable answer:} Similar to proof-based questions, some problems require sentence-form answers that are evaluated using exact string matching. This introduces high variance in rule-based verification, as even minor differences in phrasing can lead to incorrect assessments. Such variance may destabilize reinforcement learning, especially when it dominates the training signal. To mitigate this, we exclude target answers longer than 20 characters, as these are more prone to linguistic variability and harder to verify reliably.
\end{enumerate}

Through the above process, we eliminate 94\% of the candidates, yielding a curated training set of 62K math problems that are both challenging and reliably verifiable, as illustrated in Figure \ref{fig:2}. In addition, we ensure that the training data is decontaminated from common evaluation benchmarks, including AIME 2024 and 2025, and MATH 500. 

\subsection{Preliminaries of Reinforcement Learning}
\label{sec:rl_pre}

Let each data pair $(q,a)$ be \textit{i.i.d} from a distribution $\mathcal{D}$, where $q$ is a query and $a$ is the ground-truth answer. 
Given an LLM policy $\pi_{\theta}(\cdot | \cdot)$, let $o$ be an LLM-generated response to $q$, and $r(\cdot,\cdot)$ is a predefined reward function that quantifies whether the response $o$ yields $a$. 
RL-based fine-tuning aims to maximize this expected reward over $\mathcal{D}$, \textit{i.e.,} 
\begin{align}
\max_{\theta}
    J(\pi_{\theta})
    \triangleq
    \mathbb{E}_{(q,a)\sim \mathcal{D}}
    \mathbb{E}_{o\sim \pi_{\theta}(\cdot|q)}
    [r(o, a)].
    \label{eq:RL_obj}
\end{align}

\textbf{Proximal policy optimization (PPO)} \citep{PPO} is one of the most popular actor-critic RL algorithms for LLM policy optimization \citep{RLHF, OpenReasonerZero2025}. 
PPO trains both the target LLM policy $\pi_\theta$ (actor) and a value model $V_\phi$ (critic), which estimates the quality of responses generated by $\pi_\theta$. 
The PPO objective is:
\begin{equation}
\begin{aligned}
J_{\mathrm{PPO}}(\pi_\theta) \triangleq 
&\ \mathbb{E}_{(q, a) \sim \mathcal{D},\left\{o_i\right\}_{i=1}^G \sim \pi_{\theta_{\text{old}}}(\cdot \mid q)} 
\\
\Bigg[
&
\frac{1}{G} 
\sum_{i=1}^G
\frac{1}{\left| o_i \right|} 
\sum_{t=1}^{\left|o_i\right|} 
\left(
\min \Big( r_{i, t}(\theta) \textcolor{red}{\hat{A}_{i, t}(\phi)}, 
 \textcolor{red}{\operatorname{clip}}\left(r_{i, t}(\theta), 1 - \varepsilon, 1 + \varepsilon\right) \textcolor{red}{\hat{A}_{i, t}(\phi)} \Big) 
 \right)
 \Bigg],
\label{eq:PPO_obj}
\end{aligned}
\end{equation}
where $r_{i,t}(\theta)\triangleq \pi_{\theta}(o_{i,t}|q,o_{i,<t})/ \pi_{\theta_{\mathrm{old}}}(o_{i,t}|q,o_{i,<t})$ is the likelihood ratio between current and old policies; $\hat{A}_{i, t}(\phi)$ is the gated advantage estimator (GAE) estimated from $V_{\phi}(o_{i,t}|q,o_{i,<t})$. Compared to raw reward signals, GAE yields more stable policy updates \citep{OpenReasonerZero2025}.

\textbf{Group relative policy optimization (GRPO)} \citep{GRPO} is an efficient PPO variant that eliminates the critic and GAE, reducing memory and computation costs.
GRPO normalizes rewards within each rollout group to lower variance and combines likelihood ratio clipping with a KL-divergence penalty to keep $\pi_\theta$ close to the initial SFT LLM.
The GRPO objective is:
\begin{small}
\begin{equation}
\begin{aligned}
J_{\mathrm{GRPO}}(\pi_\theta) \triangleq 
&\ \mathbb{E}_{(q, a) \sim \mathcal{D},\left\{o_i\right\}_{i=1}^G \sim \pi_{\theta_{\text{old}}}(\cdot \mid q)} 
\\
\Bigg[
&
\frac{1}{G} 
\sum_{i=1}^G 
\frac{1}{\left| o_i \right|} 
\sum_{t=1}^{\left|o_i\right|} 
\left(
\min \Big( r_{i, t}(\theta) \textcolor{red}{\hat{A}_{i, t}}, 
 \operatorname{clip}\left(r_{i, t}(\theta), 1 - \varepsilon, 1 + \varepsilon\right) \textcolor{red}{\hat{A}_{i, t}} \Big) 
 -
 \textcolor{red}{\beta \mathrm{KL}(\pi_{\theta}|\pi_{\mathrm{ref}})}_{i,t}
 \right)
 \Bigg],
 \label{eq:GRPO_Obj}
\end{aligned}
\end{equation}
\end{small}
where $\hat{A}_{i,t} \triangleq (r(o_i,a) - \mathrm{mean}(\{r(o_i,a)\}_{i=1}^G))/\mathrm{std}(\{r(o_i,a)\}_{i=1}^G)$ is the group-normalized reward, and the KL term uses the K3 estimator \citep{KL_K3}. 
GRPO enables efficient large-scale LLM policy optimization, promoting long-chain CoT patterns \citep{Deepseek_R1}.

\textbf{Decoupled clip and dynamic sampling policy optimization (DAPO)} \citep{DAPO} identifies key issues in GRPO, including entropy collapse, training instability, and length bias from sample-level loss.
To address these, DAPO proposes a new objective:
\begin{equation}
\begin{aligned}
J_{\mathrm{DAPO}}(\pi_\theta) \triangleq 
&\ \mathbb{E}_{(q, a) \sim \mathcal{D},\left\{o_i\right\}_{i=1}^G \sim \pi_{\theta_{\text{old}}}(\cdot \mid q)} \\\Bigg[ 
&\quad \frac{1}{\textcolor{red}{\sum_{i=1}^G \left| o_i \right|}} \textcolor{red}{\sum_{i=1}^G \sum_{t=1}^{\left|o_i\right|}} \min \Big( r_{i, t}(\theta) \hat{A}_{i, t}, 
 \operatorname{clip}\left(r_{i, t}(\theta), 1 - \textcolor{red}{\varepsilon_{\text{low}}}, 1 + \textcolor{red}{\varepsilon_{\text{high}}}\right) \hat{A}_{i, t} \Big) \Bigg] \\
\text{subject to} \quad 
&\ \textcolor{red}{0 < \left| \left\{ o_i \,\middle|\, \text{is\_equivalent}(a, o_i) \right\} \right| < G}.
\end{aligned}
\end{equation}
DAPO decouples $\varepsilon$ into $\varepsilon_{\text{low}}$ and $\varepsilon_{\text{high}}$ to prevent entropy collapse, enforces diverse rollouts for stable gradients, and averages loss over all tokens to remove length bias. 


\subsection{Context-Aware Multi-Stage Policy Optimization (CAMPO)}
\label{sec:rl_campo}

\begin{algorithm}[t]
\caption{Context-Aware Multi-Stage Policy Optimization (CAMPO)}
\label{algo:campo}
\begin{algorithmic}[1]
\REQUIRE initial policy model $\pi_{\theta}$; reward function $r(\cdot, \cdot)$; repetition critic $f$; clip ratio models $\phi_{\text{low}}, \phi_{\text{high}}$; task prompts $\mathcal{D}$
\FOR{stage = $1, \ldots, M$}
    \STATE Update the max context length for rollout
    \STATE Sample $\varepsilon_{\text{low}}, \varepsilon_{\text{high}}$ from $\phi_{\text{low}}, \phi_{\text{high}}$
    
    \FOR{step = $1, \ldots, N$}
        \STATE Sample a batch $\mathcal{D}_b$ from $\mathcal{D}$
        \STATE Update the old policy model $\pi_{\theta_{\mathrm{old}}} \leftarrow \pi_\theta$
        \STATE Sample $G$ outputs $\{o_i\}^G_{i=1} \sim \pi_{\theta_{\mathrm{old}}}(\cdot|q)$ for each question $q \in \mathcal{D}_b$
        \STATE Filter questions where their $G$ outputs are neither all correct nor all incorrect (Equation \ref{eq:campo_objective})
        \IF{filtered questions size $< N$}
            \STATE \textbf{continue}
        \ENDIF
        \STATE For each $o_i$ of filtered questions, compute repetition penalty using $f(o_i)$, compute $\hat{A}_{i,t}$ for the $t$-th token of $o_i$ (Equation \ref{eq:campowhere})
        \FOR{iteration = $1, \ldots, \mu$}
            \STATE Update the policy model $\pi_{\theta}$ by maximizing the CAMPO objective
        \ENDFOR
    \ENDFOR
\ENDFOR
\ENSURE $\pi_\theta$
\end{algorithmic}
\end{algorithm}

We propose the context-aware multi-stage policy optimization (CAMPO) algorithm.
CAMPO enhances training by incorporating awareness of both {context length} and {content}.
Specifically, multi-stage training with gradually increasing context length improves efficiency, while a repetition penalty encourages redundancy-aware learning and promotes more effective reasoning. Training progresses to the next stage once the response length saturates.
The CAMPO training objective is formalized as:
\begin{equation}
\begin{aligned}
J_{\mathrm{CAMPO}}(\pi_\theta) \triangleq 
&\ \mathbb{E}_{(q, a) \sim \mathcal{D},\left\{o_i\right\}_{i=1}^G \sim \pi_{\theta_{\text{old}}}(\cdot \mid q)} \\\Bigg[ 
&\quad \frac{1}{{\sum_{i=1}^G \left| o_i \right|}} {\sum_{i=1}^G \sum_{t=1}^{\left|o_i\right|}} \min \Big( r_{i, t}(\theta) \hat{A}_{i, t}, 
 \operatorname{clip}\left(r_{i, t}(\theta), 1 - \textcolor{red}{\phi_{\text{low}}(s)}, 1 + \textcolor{red}{\phi_{\text{high}}(s)}\right) \hat{A}_{i, t} \Big) \Bigg] \\
\text{subject to} \quad 
&\ {0 < \left| \left\{ o_i \,\middle|\, \text{is\_equivalent}(a, o_i) \right\} \right| < G},
\end{aligned}
\label{eq:campo_objective}
\end{equation}
where
\begin{small}
\begin{equation}
    r_{i,t}(\theta)\triangleq \frac{\pi_{\theta}(o_{i,t}|q,o_{i,<t})}{\pi_{\theta_{\mathrm{old}}}(o_{i,t}|q,o_{i,<t})},\quad
    \hat{A}_{i,t} \triangleq \frac{r(o_i,a) - \textcolor{red}{{f(o_i)}} - \mathrm{mean}(\{r(o_i,a) - {f(o_i)}\}_{i=1}^G)}{\mathrm{std}(\{r(o_i,a) - \textcolor{red}{f(o_i)}\}_{i=1}^G)},
\label{eq:campowhere}
\end{equation}
\end{small}
$s$ denotes the training stage, $\phi_{\text{low}}(s)$ and $\phi_{\text{high}}(s)$ represent the corresponding decoupled clipping distributions for different stages.
$f(o_i)$ measures repetition in $o_i$ and outputs a score between 0 and 1, depending on how early the repetition occurs.
Specifically, $f(o_i)$ returns the proportion of the whole token sequence that lies inside the detected repeating loop.
The full algorithm can be found in Algorithm \ref{algo:campo}.

\subsubsection{Efficiency-Aware Perspective on Multi-Stage Training}


One of the main bottlenecks in RL training lies in the high computational cost of long rollouts, which can significantly slow down training and reduce sampling efficiency.
To address this, inspired by DeepScaleR \citep{deepscaler2025} and Skywork-OR1 \citep{skywork-or1-2025}, we adopt a multi-stage RL training strategy designed to enhance training efficiency and model performance.
In this framework, the maximum permitted response length is progressively increased across training stages, as illustrated in Algorithm \ref{algo:campo}.
As rollouts exceed the permitted response will be assigned zero reward, it is necessary to increase the lengths to allow more complex reasoning during training.
\begin{wrapfigure}{r}{0.51\textwidth}
        \centering
        \includegraphics[width=0.51\textwidth]{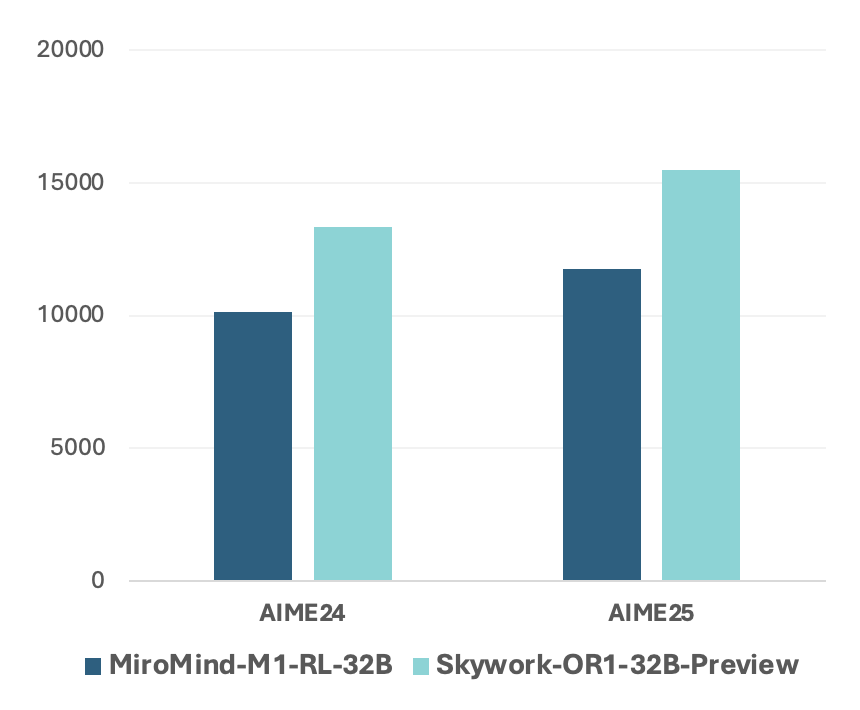}
        \caption{Average token length of model responses computed across all rollout attempts, evaluated over 64 independent runs on the corresponding test sets. This analysis includes both correct and incorrect answers, highlighting overall response efficiency regardless of correctness. 
        }\label{fig:mean_response_len}
        \vspace{-3em}
\end{wrapfigure}

This staged setup offers several benefits. First, the shorter response limit helps constrain the model's output space, reducing rollout length and accelerating feedback cycles.
Furthermore, responses that exceed the current maximum length are treated as failures, providing a clear training signal that encourages the model to produce more concise and refined outputs. This effect is demonstrated in Figure \ref{fig:mean_response_len}, which shows the mean token length of model outputs over 64 runs on AIME24 and AIME25 benchmarks. Compared to \texttt{Skywork-OR1-32B-Preview}, \texttt{MiroMind-M1-RL-32B} generates significantly shorter responses with minimal impact on accuracy, as shown in Table \ref{tab:7b_table_comparison}.

As training advances, progressively relaxing the length constraint becomes essential, especially for complex tasks that inherently demand longer reasoning trajectories.
This gradual increase allows the model to scale its capacity for deeper reasoning while preserving the efficiency gains from earlier stages. 

\subsubsection{Stabilizing Training with Repetition Penalty}

The training process of RL can be unstable and prone to sudden collapse, particularly when the model starts generating a narrow set of low-diversity outputs. 
\begin{wrapfigure}{l}{0.45\textwidth}
    \centering
    \includegraphics[width=0.45\textwidth]{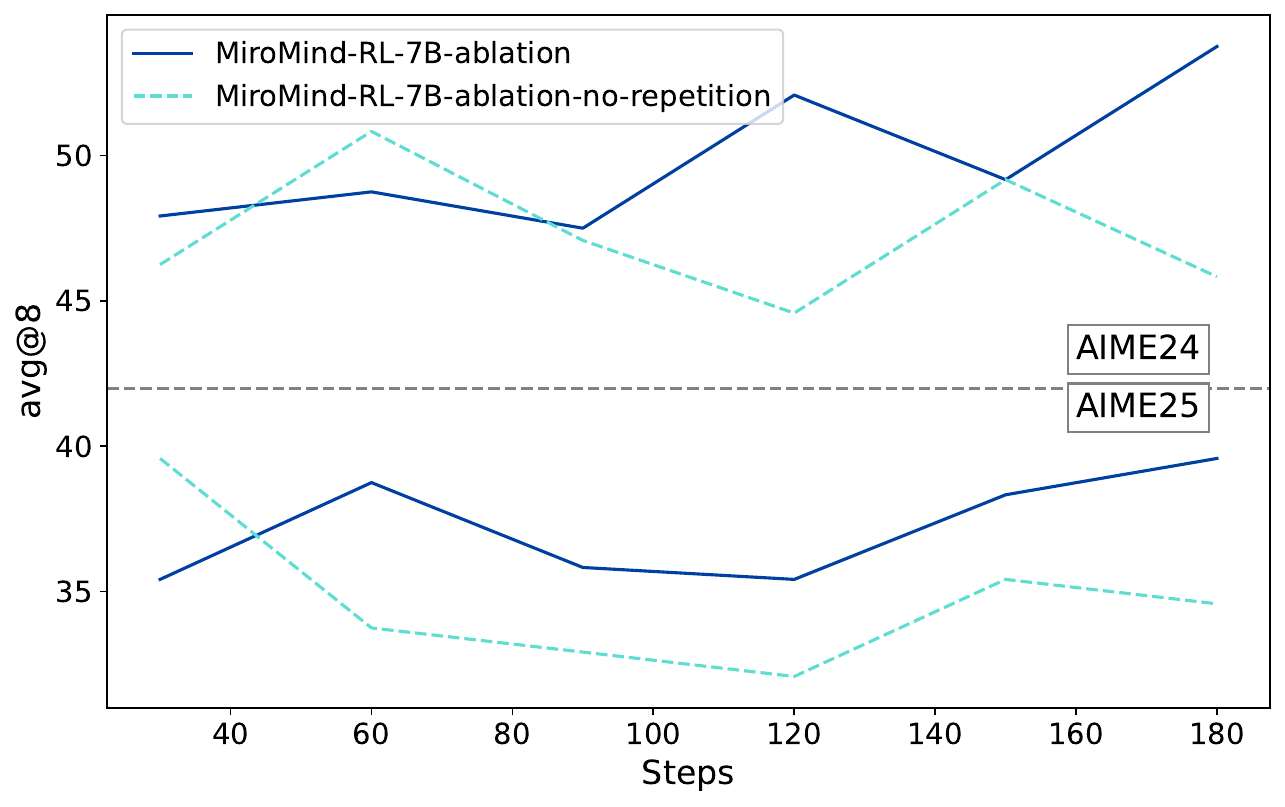}
      \caption{The training process is more stable with repetition penalty. 
      }
      \label{fig:ablation_rep}
\end{wrapfigure}
To encourage the emergence of long reasoning patterns, the KL loss is often omitted; however, this can cause the trained model to drift significantly from the original policy model, potentially resulting in suboptimal behaviors.
Introducing a repetition penalty helps mitigate this issue by promoting output diversity and discouraging repetitive patterns, thereby preventing the policy from collapsing into a narrow output space.
As described in Section \ref{sec:rl_campo} and Algorithm \ref{algo:campo},
the repetition score ranges from 0 to 1, depending on how early the repetition occurs. Earlier repetitions incur heavier penalties. 
This design allows the penalty to dynamically adapt to the model's repetition behavior. 
Figure \ref{fig:ablation_rep} presents the training performance on AIME24 and AIME25 for CAMPO with the 7B model, compared to CAMPO without the repetition penalty.
As shown, incorporating the repetition penalty leads to a more stable training process.

\subsubsection{Accurate Verifier for Efficient Reasoning}

The success of RL in training language models depends heavily on the accuracy of the reward signals, which act as the model’s very few forms of feedback. 
Unlike supervised learning, where models are guided by explicit ground-truth labels, RL relies on these signals to shape behavior indirectly. If the reward function mistakenly penalizes correct answers, often due to verifier errors, it can lead the model to develop inefficient reasoning patterns.

        \begin{figure}[htb]
            \begin{minipage}[t]{.51\textwidth}
                \centering
                \includegraphics[width=\textwidth]{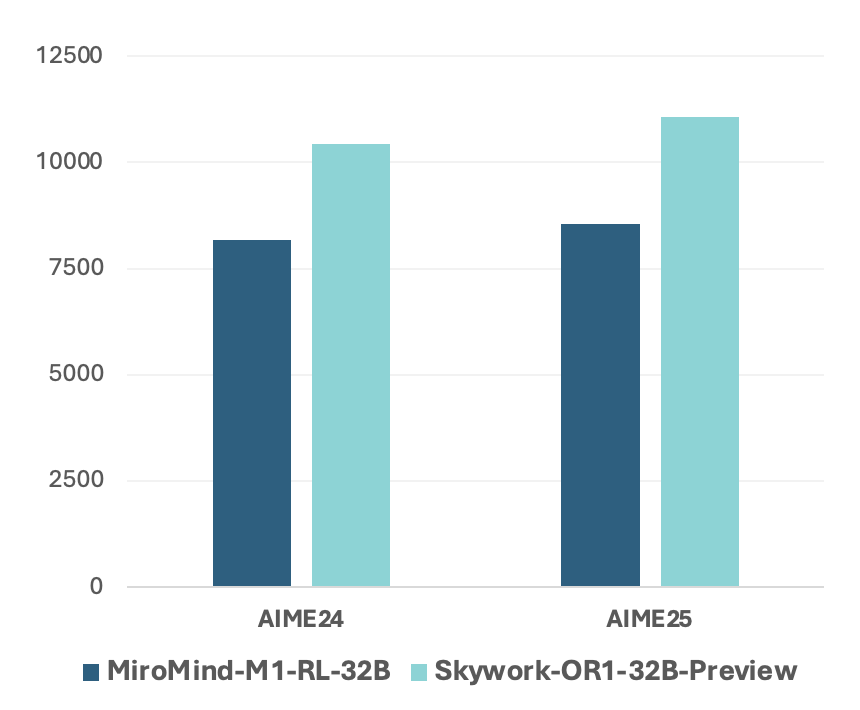}
        \caption{Average token count of model responses conditioned on correct answers, computed over all rollouts and averaged across 64 runs on the respective test sets. Unlike Figure~\ref{fig:mean_response_len}, this analysis considers only responses that received correct reward signals, highlighting that correct, rewarded outputs appear more efficient. 
        }\label{fig:correct_response_len}
            \end{minipage}
            \hfill 
            \begin{minipage}[t]{.47\textwidth}
                \centering
                \includegraphics[width=\textwidth]{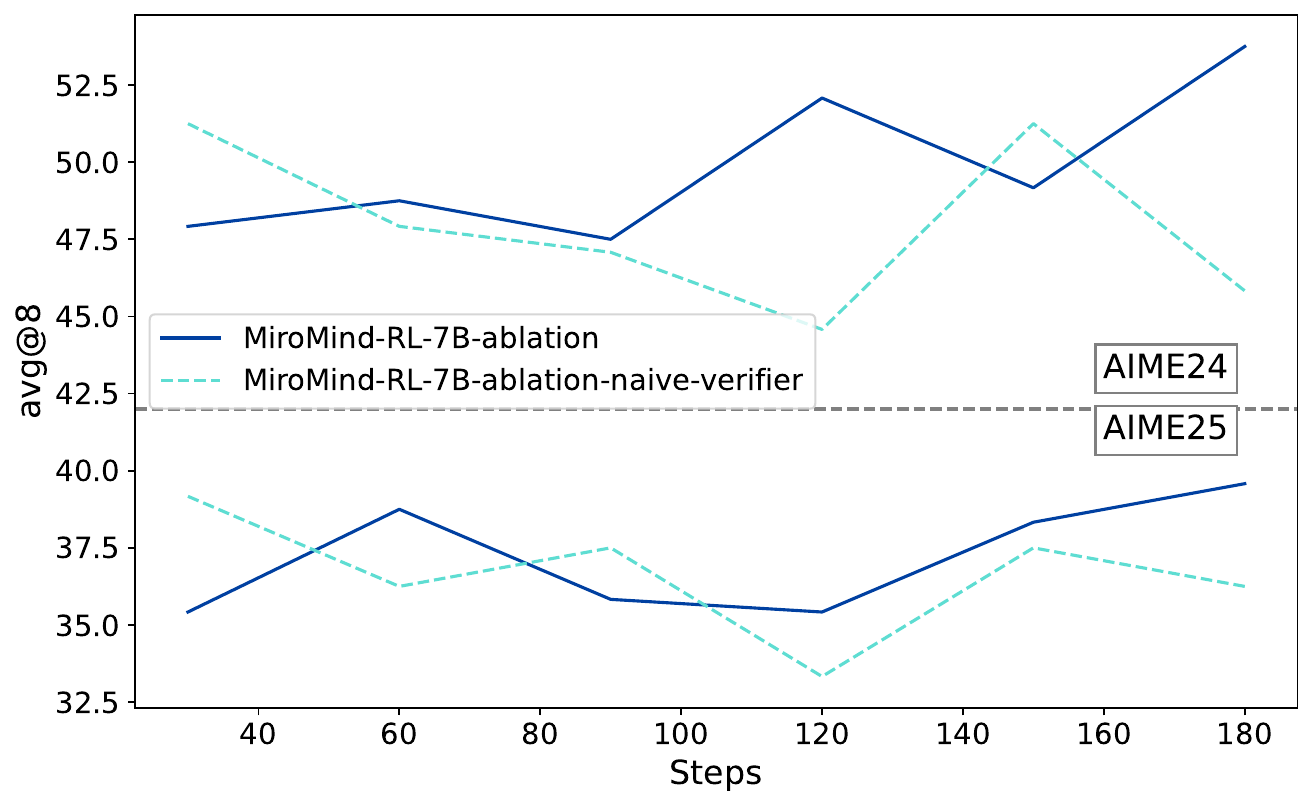}
                  \caption{RL benefits from a more accurate verifier, highlighting the critical role of verifier quality in enabling robust training. High-quality verification ensures that reward signals align more closely with ground truth correctness, reducing noise and inconsistencies during optimization.}
                  \label{fig:ablation_verifier}
            \end{minipage}
        \end{figure}

This problem becomes even more important in settings like GRPO, where the model’s internal state is estimated based on the outcomes of a batch of rollouts. 
{Inaccurate reward signals, such as when correct answers are penalized due to verifier failure, can distort the model's belief about which behaviors are desirable. As a result, it may revise valid solutions or introduce unnecessary reasoning steps that offer little value.}
Over time, this misalignment can cause the model to adopt overly complicated or even unstable reasoning behaviors \citep{zeng2025simplerlzooinvestigatingtamingzero}.

To reduce these issues, we significantly improved the math verifier \footnote{\href{https://github.com/huggingface/Math-Verify}{https://github.com/huggingface/Math-Verify}}. The original version (v0.7.0) struggled with various edge cases, including answers involving units, constants like $\pi$ and degrees, percentages ($\%$), and small differences in numerical precision. Our updated verifier introduces a cascade design with multiple verification stages that gradually increase in strictness. We also incorporated a range of human-curated fixes to make the verifier more robust and comprehensive. Together, these changes lead to more reliable feedback during training, helping the model learn to generate concise and logically sound answers more efficiently.

            
The impact of these improvements is illustrated in Figure \ref{fig:correct_response_len}, which shows the average token count of correct model responses across 64 runs on AIME24 and AIME25. Compared to \texttt{Skywork-OR1-32B-Preview}, \texttt{MiroMind-M1-RL-32B} consistently produces significantly shorter and more efficient reasoning traces for responses that received correct reward signals.
As further shown in Figure \ref{fig:ablation_verifier}, RL benefits from more accurate verification, underscoring the importance of verifier quality in shaping robust training.


\subsection{MiroMind-M1-RL-32B}
\label{sec:rl_32b}

In this section, we present \texttt{MiroMind-M1-RL-32B}, our direct attempt with CAMPO on SOTA reasoning models.

\subsubsection{Training Details} 


To demonstrate the effectiveness of our RL settings with CAMPO on state-of-the-art models, we initialize training from \texttt{DeepSeek-R1-Distill-Qwen-32B} using the dataset described in Section~\ref{sec:datapre}.
For multi-stage training, we begin with a maximum response length of 16,384 in Stage 1, then progressively increase it to 32,768 and 49,152 in Stages 2 and 3, respectively. 
All other training configurations remain consistent across stages.
Specifically, we omit the KL loss to allow more flexibility in model learning. 
We use a constant learning rate of 1e-6 and a clipping ratio of 0.2. 
During rollouts, the temperature is set to 1.0 for more diverse sampling, and each data sample is rolled out 16 times. 
Both the training and mini-batch sizes are set to 32, ensuring strictly on-policy rollouts that better reflect the model’s current state.
All experiments on \texttt{DeepSeek-R1-Distill-Qwen-32B} are conducted using 16*8 A100 GPUs.
We follow the same evaluation setting as described in Section \ref{sec:SFT}, except that the temperature is set to 1.0 to match the setting used during RL training.

\subsubsection{Experimental Results and Insights}

    \begin{table}[t!]
    \centering
    \resizebox{0.65\textwidth}{!}{
    \begin{tabular}{lrrr}
    \toprule
    \textbf{Model} & \textbf{AIME24} & \textbf{AIME25} & \textbf{MATH500} \\
    \midrule
    DeepSeek-R1 & 79.8 & 70.0 & --  \\
    DeepSeek-R1-0528 & 91.4 & 87.5 & --  \\
    Qwen3-8B & 76.0 & 67.3 & -- \\
    DeepSeek-R1-0528-Qwen3-8B & 86.0 & 76.3 & --   \\
    MiMo-7B-RL & 68.2 & 55.4 & 95.8  \\

    \midrule
    \multicolumn{4}{l}{\emph{32B Models trained from Qwen2.5 series}} \\
    \midrule
    DeepSeek-R1-Distill-Qwen-32B & 70.8 & 52.1 & 95.8  \\
    Skywork-OR1-32B-Preview & 77.1 & 68.2 & 97.5  \\
    \cellcolor{custombluedark!30}MiroMind-M1-RL-32B & \cellcolor{custombluedark!30}77.5 & \cellcolor{custombluedark!30}65.6 & \cellcolor{custombluedark!30}96.4 \\
    \midrule
    \multicolumn{4}{l}{\emph{7B Models trained from Qwen2.5 series}} \\
    \midrule
    DeepSeek-R1-Distill-Qwen-7B & 55.5 & 39.2 & --  \\
    MiroMind-M1-SFT-7B & 60.4 & 45.0 & 94.6  \\
    Light-R1-7B-DS & 59.1 & 44.3 & --  \\
    Skywork-OR1-7B & 72.2 & 54.6 & --  \\
    \cellcolor{custombluedark!30}MiroMind-M1-RL-7B & \cellcolor{custombluedark!30}73.4 & \cellcolor{custombluedark!30}57.8 & \cellcolor{custombluedark!30}96.7 \\

    \bottomrule
    \end{tabular}
    }
    \caption{Comparison of our 7B and 32B model performance across benchmark datasets.}
    \label{tab:7b_table_comparison}
    \end{table}

    \begin{figure}[htbp]
      \centering
      \begin{subcaptionbox}{AIME24\label{fig:sub1}}[0.45\textwidth]
        {\includegraphics[width=\linewidth]{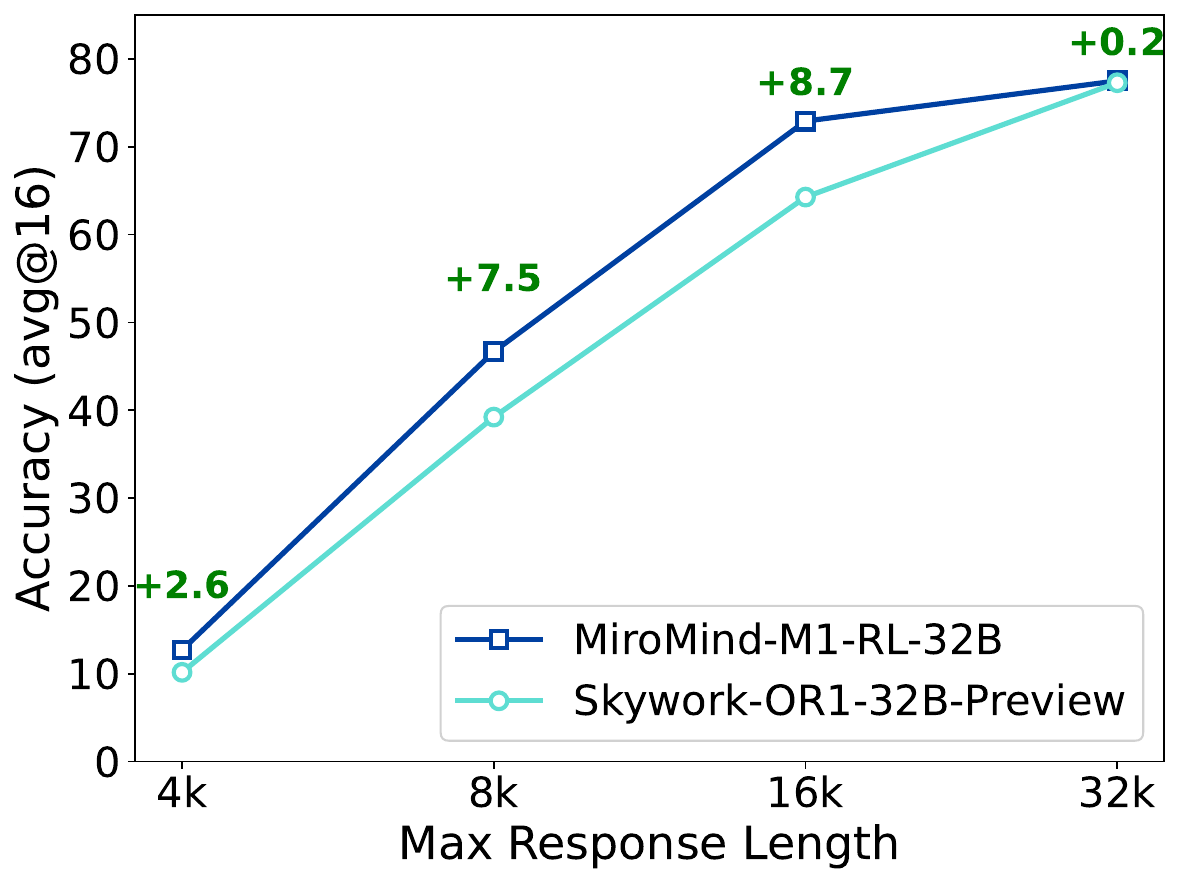}}
      \end{subcaptionbox}
      \hspace{0.05\textwidth}
      \begin{subcaptionbox}{AIME25\label{fig:sub2}}[0.45\textwidth]
        {\includegraphics[width=\linewidth]{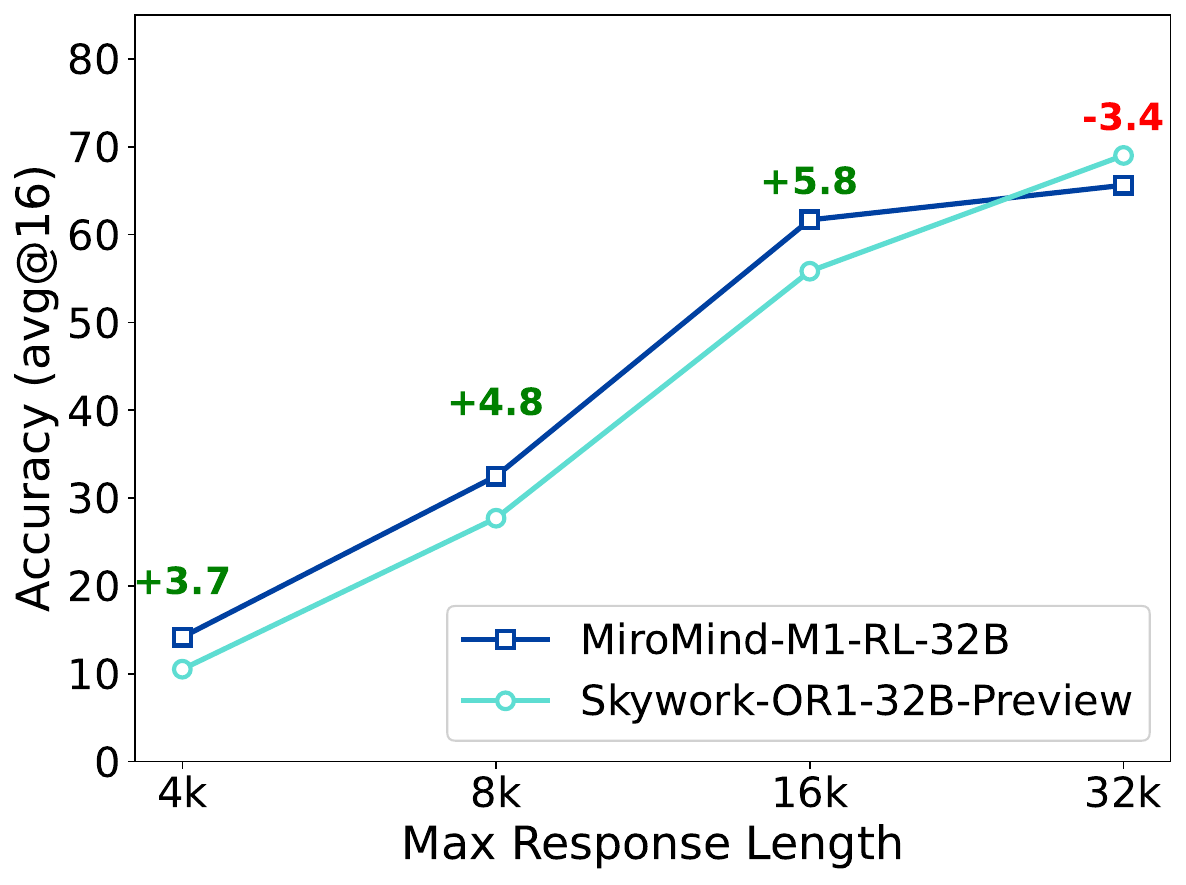}}
      \end{subcaptionbox}
      \caption{Comparison of \texttt{MiroMind-M1-RL-32B} and \texttt{Skywork-OR1-32B-Preview} on AIME24 and AIME25 across different maximum response lengths.}
      \label{fig:combined}
    \end{figure}

As shown in Table \ref{tab:7b_table_comparison}, \texttt{MiroMind-M1-RL-32B} demonstrates that CAMPO can significantly enhance performance on math benchmarks when finetuned from a strong initialization checkpoint (\textit{i.e.,} \texttt{DeepSeek-R1-Distill-Qwen-32B}).
Specifically, it achieves notable improvements of 6.7\% on AIME24 and 13.5\% on AIME25, indicating the effectiveness of our RL framework.

Nevertheless, our 32B model still lags behind certain SOTA models that also employ RL for training.
For instance, \texttt{Skywork-OR1-32B-Preview} surpasses our model by 2.6\% on AIME25.
We attribute this performance gap primarily to differences in the composition of training data.
\texttt{Skywork-OR1-32B-Preview} benefits from a diverse mixture of math and code data, where the inclusion of code likely helps the model develop stronger symbolic reasoning capabilities.
In contrast, our RL training relies solely on math-focused data, which may limit the model's generalization on certain problem types.

On the other hand, CAMPO leads to more token-efficient reasoning behavior.
As illustrated in Figure \ref{fig:combined}, \texttt{MiroMind-M1-RL-32B} consistently outperforms \texttt{Skywork-OR1-32B-Preview} across all shorter max response lengths, ranging from 4k to 16k.
This suggests that our model is able to arrive at correct answers more succinctly.
We attribute this token-efficiency to three key components in CAMPO:
the repetition penalty, the cascade verifier, and the multi-stage training strategy.
Together, these components reduce redundant reasoning and encourage the model towards generating token-efficient yet accurate responses, rather than unnecessarily prolonging outputs.

\subsection{MiroMind-M1-RL-7B}
\label{sec:rl_7b}

    In this section, we present \texttt{MiroMind-M1-RL-7B}, our strongest model derived from the \texttt{Qwen2.5} pretrained base series. This model represents a fully transparent and reproducible workflow for building reasoning-capable models through supervised fine-tuning (SFT) followed by reinforcement learning (RL). Starting from the \texttt{Qwen2.5-7B-Math-Base} checkpoint, the model is first fine-tuned via SFT (as introduced in Section~\ref{sec:SFT}) to reach \texttt{MiroMind-M1-SFT-7B}, then further optimized with RL to enhance its mathematical reasoning capabilities. As shown in Table~\ref{tab:7b_table_comparison}, it achieves the best performance among all Qwen2.5-based models across multiple benchmarks. Further implementation details, comparative analysis, and insights from reinforcement learning are presented in the sections below to support reproducibility and highlight directions for future improvement.

    \subsubsection{Training Details}

    First, we use the same set of training data as the 32B model for RL training. All data points are accompanied by verifiable outputs. While it is commonly argued that the difficulty level of RL training data plays a crucial role—particularly in early training stages where smoother reasoning progression is preferred. We find that it is sufficient for the dataset to span a range of difficulties for our algorithms. Notably, the 32B training data does not negatively impact the training of our 7B model. We attribute this to two key factors:
    1. In our algorithm design, we explicitly exclude rollout samples that are either entirely correct or entirely incorrect, retaining only partially correct samples for training, as described in Algorithm~\ref{algo:campo};
    2. During training, we ensure that each batch contains a sufficient number of valid samples (i.e., samples with partially correct rollouts) before performing gradient updates, maintaining a stable and efficient optimization process.

    Second, for initial checkpoint, we use our post-trained reasoning model \texttt{MiroMind-M1-SFT-7B} as the initial checkout which is shown to be a better reasoning model compared to \texttt{DeepSeek-R1-Distill-Qwen-7B} and is fully reproducible with open-source data.

    Third, similar to our 32B model variant, we adopt multiple training stages with varying maximum generation lengths. This strategy primarily aims to improve training efficiency, as the rollout phase dominates the computational cost in RL training, and generation length is a critical factor influencing that cost. Further details on the multi-stage setup and comparisons are provided in the following sections.

    All experiments are conducted on a distributed setup consisting of 8 compute nodes, each equipped with 8 NVIDIA A800 GPUs, totaling 64 GPUs. The training pipeline is implemented using the VERL\footnote{\url{https://github.com/volcengine/verl}} framework, which provides efficient infrastructure support for large-scale reinforcement learning workflows.

    
        \begin{figure}[htb]
          \centering
          \includegraphics[width=0.7\textwidth]{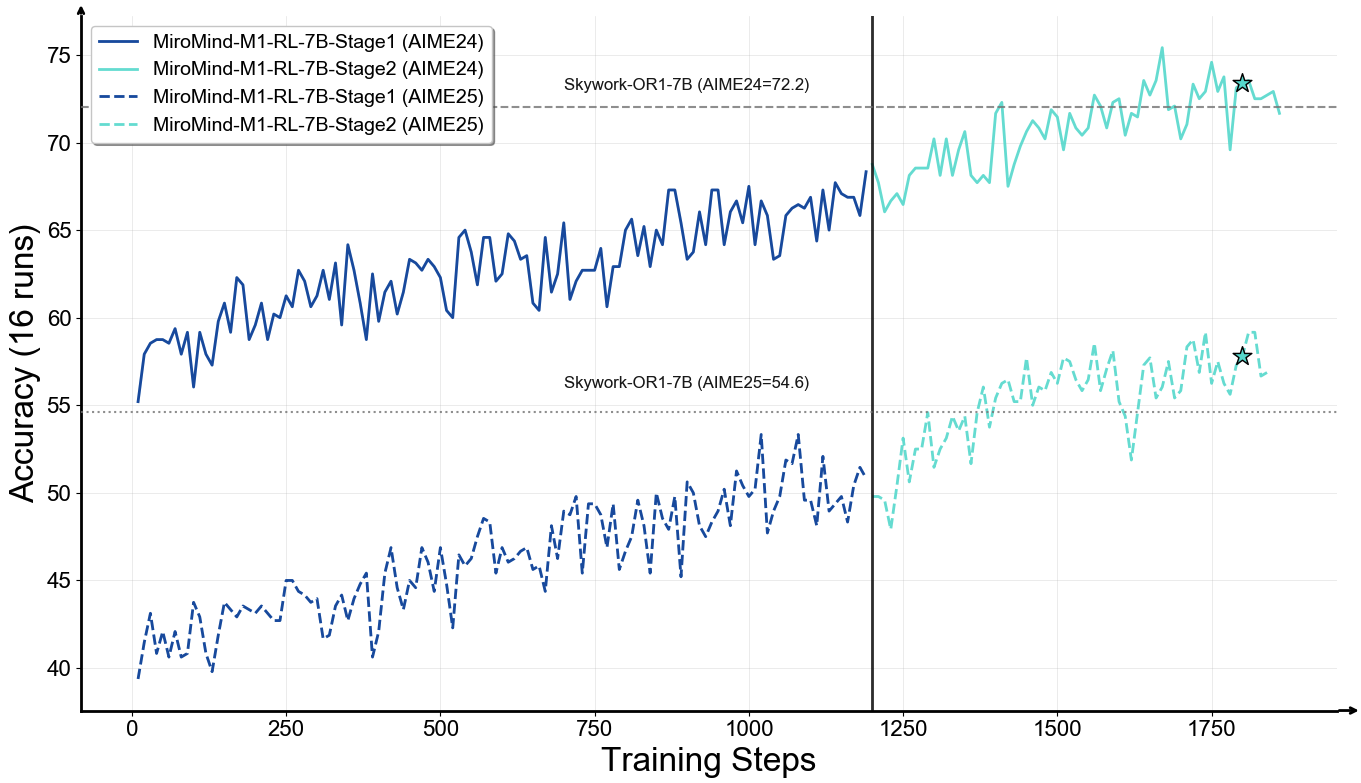}
          \caption{The model's performance steadily improves throughout the training process.}
          \label{fig:7b_performance_training}
        \end{figure}

    \subsubsection{Experimental Results and Insights}
        
        We adopt a two-stage training strategy. In the first stage, we cap the maximum rollout length at 16K tokens, discarding samples that fail to produce a complete answer within this limit. This constraint significantly reduces training cost, as the rollout phase dominates computational overhead and scales with sequence length. After 1200 steps, we update the maximum rollout length per sample to 32K tokens and continue training with the same model checkpoint and optimizers. We summarize notable patterns observed during training and analyze their impact on model performance and optimization dynamics.

        \textbf{Model Performance Over Training Steps}. In Figure~\ref{fig:7b_performance_training}, we show the performance trajectory of our \texttt{MiroMind-M1-RL-7B} model during training. The RL process yields over a 15\% accuracy improvement on both AIME24 and AIME25. As a result, our model achieves state-of-the-art (SOTA) math capabilities among all models derived from the \texttt{Qwen2.5} series. 
        While some recently introduced models outperform ours, they are built upon newer base models. Our techniques can be similarly applied to those newer models if needed. Additionally, our model outperforms \texttt{MiMo-7B-RL}~\citep{xiaomi2025mimo}, which is trained from scratch using 500k SFT samples, a training size comparable to ours. A newer variant, \texttt{MiMo-7B-RL-0530}, was recently released with 6 million newly curated SFT samples, making direct comparison less appropriate due to the substantially larger training set.


        \begin{figure}[htb]
          \centering
          \includegraphics[width=0.7\textwidth]{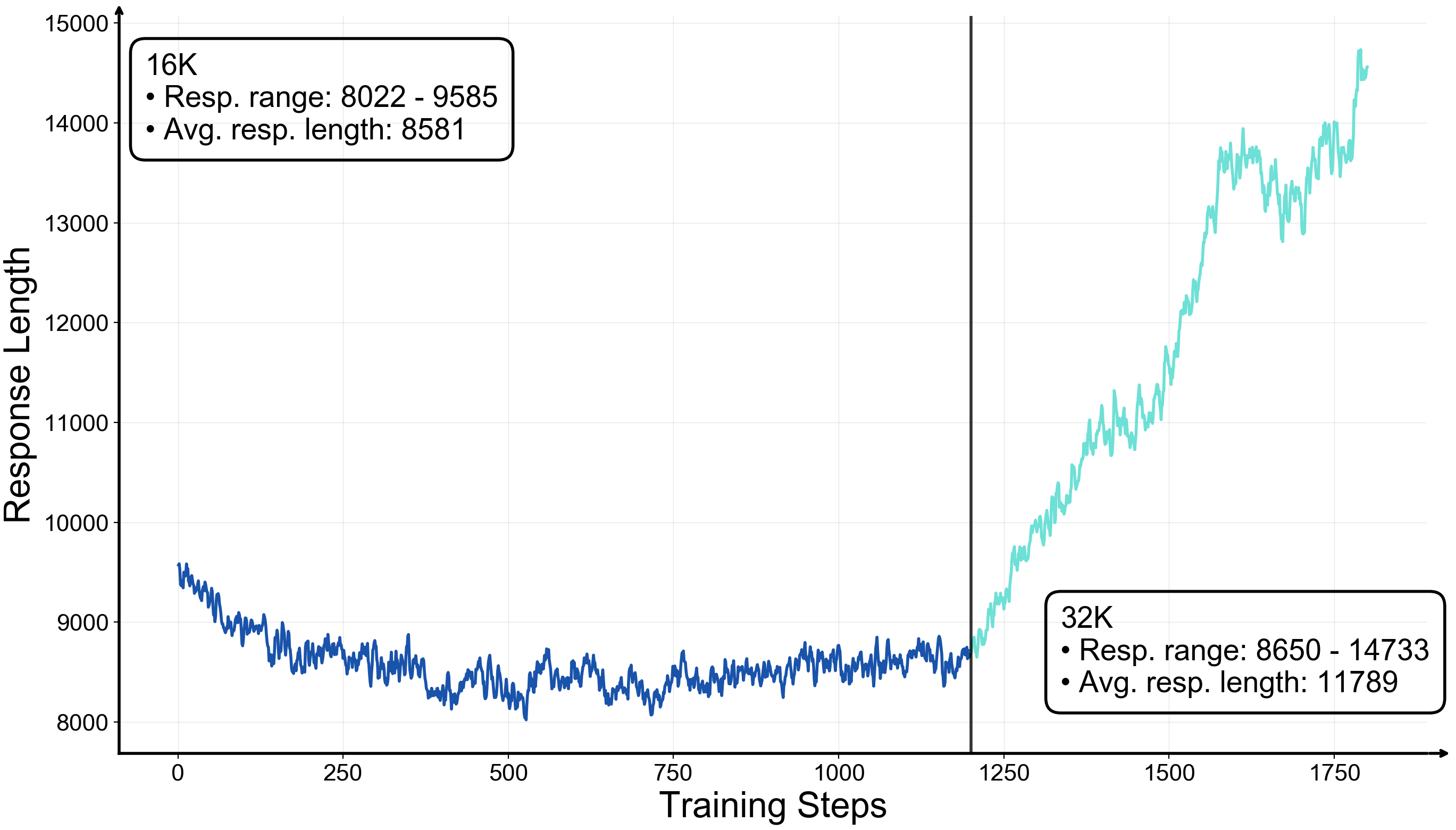}
          \caption{Response length trend during two-stage training. Under the 16K generation cap, the model generates relatively shorter responses. Once the limit is increased to 32K, the model begins producing significantly longer outputs.}
          \label{fig:response_length_7b}
        \end{figure}

        \textbf{Response Length}. We examine the average generation length across all training samples throughout the training process. As shown in Figure~\ref{fig:response_length_7b}, the response length becomes compressed during the initial phase with a 16K token generation cap, stabilizing between 8K and 9K tokens. After transitioning to the 32K generation limit, the average maximum length increases significantly, with responses extending beyond 13K tokens. We hypothesize that the initial constraint pushes the model to operate near its capacity within a limited budget, helping establish a strong foundation for reasoning. This, in turn, may lead to more stable and effective optimization in the longer sequence regime.

        \begin{figure}[htb]
          \centering
          \includegraphics[width=0.75\textwidth]{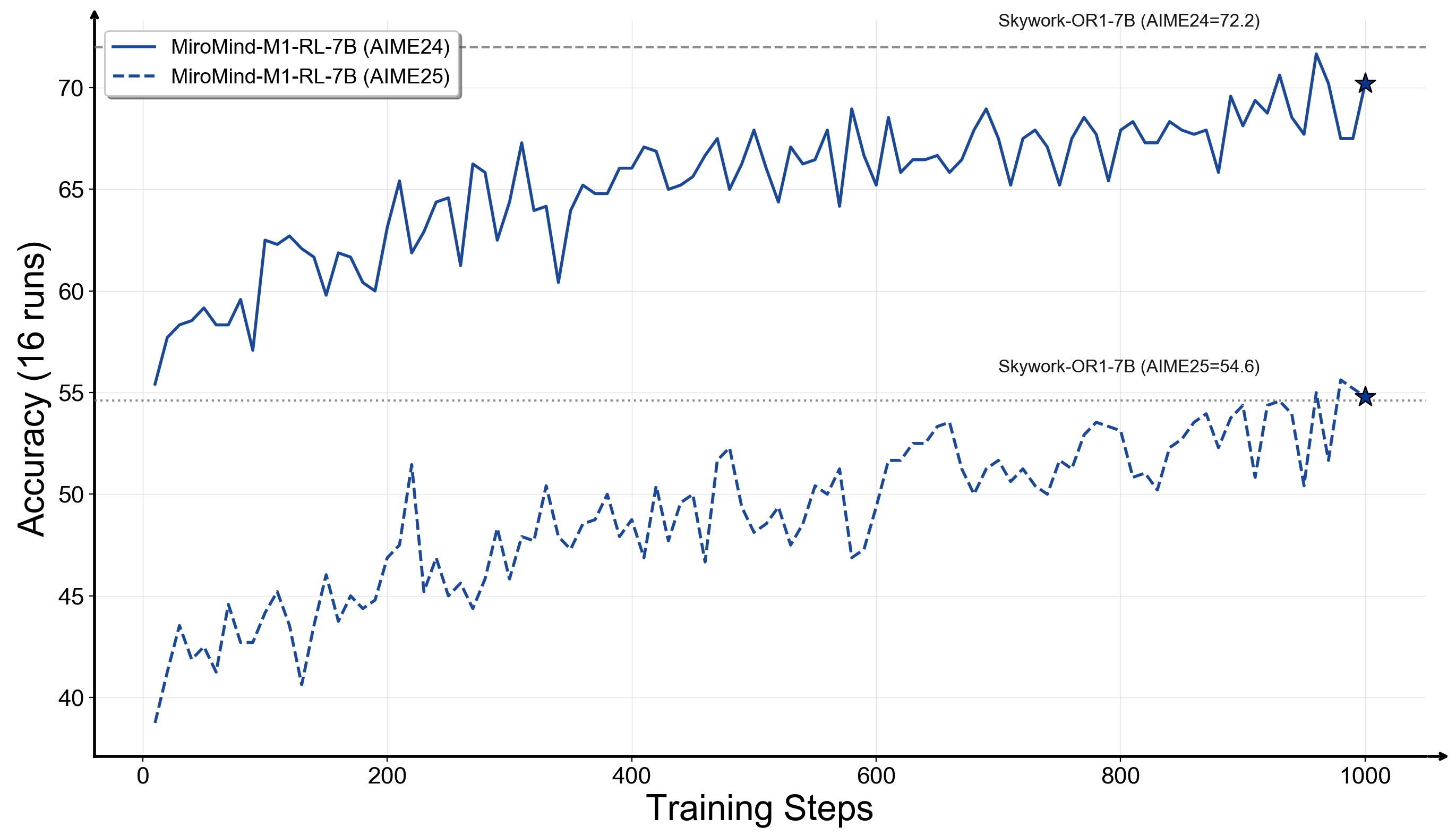}
          \caption{Performance trend of the model trained using a single-stage 32K max context length schema.}
          \label{fig:7b_one_stage}
        \end{figure}

        \textbf{Comparison with Single-Stage Training}. As a parallel study to our two-stage pipeline, we experiment using a single-stage training process with the 7B model by setting the maximum context length to 32K. All hyperparameters and settings were kept consistent with those used in \texttt{MiroMind-M1-RL-7B}. The full training performance trajectory is shown in Figure~\ref{fig:7b_one_stage}. While the single-stage approach can achieve competitive performance compared to the two-stage method, the latter offers two key advantages: (1) faster training: since starting with long sequences imposes greater memory and computational overhead, resulting in slower training speeds; and (2) potentially better performance when the output token budget is limited, as same patterns found in our \texttt{MiroMind-M1-RL-32B} models, which also experienced multi-stage training, presented in Figure~\ref{fig:combined}.

        \begin{figure}[htb]
          \centering
          \includegraphics[width=0.95\textwidth]{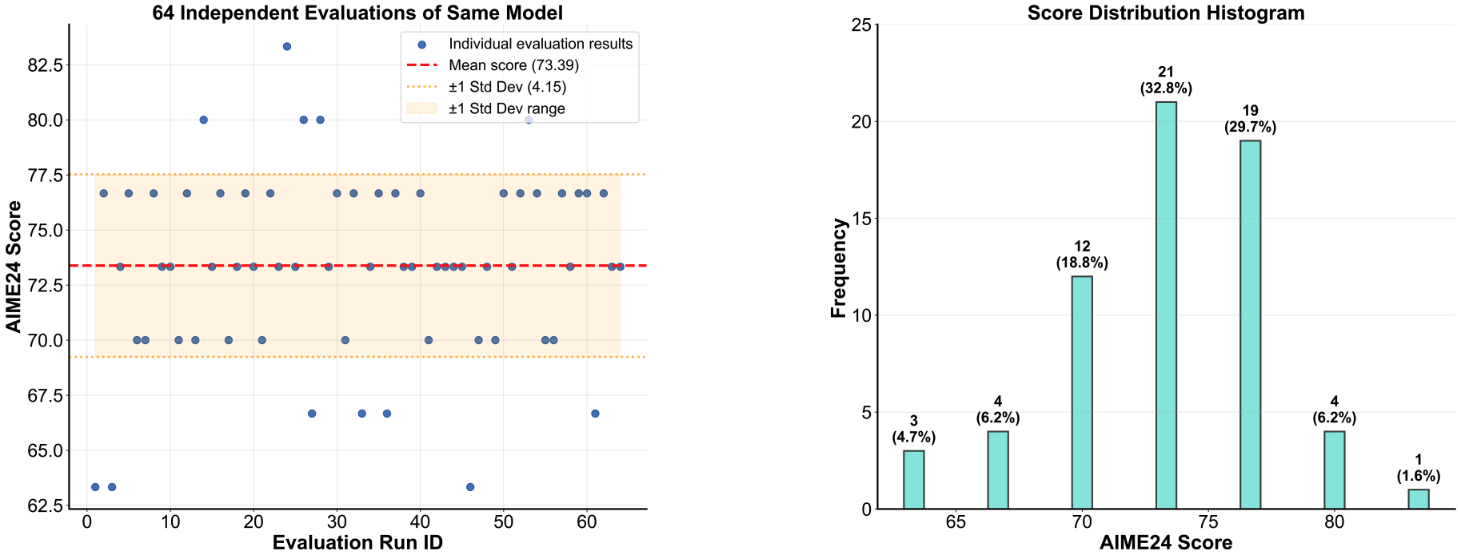}
          \caption{Evaluation stability assessment: 64 repeated evaluations of \texttt{MiroMind-M1-RL-7B} model performance on AIME24.}
          \label{fig:robust_evaluation}
        \end{figure}

        \textbf{Discussion on Evluation Stability}. There are growing concerns about the stability of evaluation results, particularly for challenging benchmarks like AIME24 and AIME25, which contain only 30 questions. A difference of just 1–2 correct answers can cause performance fluctuations exceeding 5\%, posing significant challenges for consistent benchmarking. A common mitigation strategy is to run the evaluation 64 times and report the average accuracy. However, as shown in Figure~\ref{fig:robust_evaluation}, the two-sided standard deviation can still exceed 8\%. While we currently lack a better solution, increasing the number of evaluation runs can yield more robust results, but at the cost of a significantly more time-consuming benchmarking process.       

        \textbf{Efficiency in RL Training}. Another important aspect to consider is the efficiency of reinforcement learning (RL) training. We found that the primary bottleneck lies not in updating model parameters, but in the rollout phase, where the model generates responses to compute rewards. In our current framework, rollouts are executed in a synchronized manner across batches. As a result, the presence of even a few samples with extremely long generations can significantly delay the entire batch, leading to prolonged GPU idle time and reduced overall training efficiency. This inefficiency becomes more pronounced when encountering long-tail generation issues during inference, where certain inputs result in abnormally long outputs. Such cases dominate the runtime and disrupt the parallelism benefits typically expected in batch processing. Similar challenges have been reported in prior work~\cite{AM}, where researchers proposed various strategies such as detached rollout and streaming load balancing architecture. Our two-stage training strategy partially alleviates this issue by bounding the generation length in the early stage, thus containing rollout variability to some extent. However, addressing this bottleneck more comprehensively will require further research and system-level optimizations in future work.

\section{Conclusions}
In this paper, we open-source a comprehensive system for training reasoning language models, including codes, datasets, and checkpoints.
For the SFT phase, we curate a large scale of high-quality data, which results in significantly stronger performance on math reasoning benchmarks compared to DeepSeek's 7B distillation models.
In addition, we propose CAMPO, a context-aware multi-stage policy optimization framework, to further improve the performance of the reinforcement learning stage.
We show that our reasoning models of 7B can achieve better performance than Skywork’s counterpart with fewer tokens, indicating greater efficiency.
We hope that our efforts can facilitate future research on reasoning language models.

\bibliography{reference}
\bibliographystyle{reference}

\clearpage

\end{document}